\pdfoutput=1

\documentclass[11pt]{article}

\usepackage[final]{acl}
\usepackage{times}
\usepackage{latexsym}

\usepackage[T1]{fontenc}

\usepackage[utf8]{inputenc}

\usepackage{microtype}

\usepackage{inconsolata}

\usepackage{graphicx}
\usepackage{booktabs} 

\usepackage[scr=rsfs]{mathalpha}
\usepackage{subcaption}
\usepackage{array}

\usepackage{amsfonts}
\usepackage{color}
\usepackage{amsmath}
\usepackage{multirow}
\usepackage{enumitem}

\title{Bridging the Visual Gap: Fine-Tuning Multimodal Models \\ with Knowledge-Adapted Captions}

\author{
    Moran Yanuka\textsuperscript{\(\tau\)} \:\:\:\: Assaf Ben-Kish\textsuperscript{\(\tau\)} \:\:\:\: Yonatan Bitton\textsuperscript{\(G\)} \:\:\:\: Idan Szpektor\textsuperscript{\(G\)} \:\:\:\: Raja Giryes\textsuperscript{\(\tau\)} \\
    \\
    \textsuperscript{\(\tau\)}Tel Aviv University \:\:\:\:\: \textsuperscript{\(G\)}Google Research
}

\newcommand{\methodname}{KnowAda}

\definecolor{bubblegum}{rgb}{0.99, 0.76, 0.8}
\definecolor{corn}{rgb}{0.98, 0.93, 0.36}
\definecolor{cream}{rgb}{1.0, 0.99, 0.82}
\definecolor{bluebell}{rgb}{0.64, 0.64, 0.82}
\definecolor{brilliantlavender}{rgb}{0.96, 0.73, 1.0}
\definecolor{brightube}{rgb}{0.82, 0.62, 0.91}
\definecolor{lavenderindigo}{rgb}{0.58, 0.34, 0.92}
\definecolor{lemonchiffon}{rgb}{1.0, 0.98, 0.8}
\definecolor{lightblue}{rgb}{0.68, 0.85, 0.9}
\definecolor{lightgreen}{rgb}{0.67, 0.88, 0.69}
\definecolor{lightred}{rgb}{0.99, 0.5, 0.5}
\definecolor{awesome}{rgb}{1.0, 0.13, 0.32}

\definecolor{lighterred}{rgb}{0.99, 0.7, 0.7}

\definecolor{darkgreen}{rgb}{0.0, 0.6, 0.0}
\definecolor{darkred}{rgb}{0.7, 0.0, 0.0}

\begin{document}
\maketitle

\begin{abstract}

Recent research increasingly focuses on training vision-language models (VLMs) with long, detailed image captions. However, small-scale VLMs often struggle to balance the richness of these captions with the risk of hallucinating content during fine-tuning. In this paper, we explore how well VLMs adapt to such captions. To quantify caption quality, we propose Decomposed NLI (DNLI), an evaluation framework that breaks down generated captions into individual propositions, assessing each in isolation. This fine-grained analysis reveals a critical balance between capturing descriptive details and preventing hallucinations. Our findings show that simply reducing caption complexity or employing standard data curation techniques does not effectively resolve this issue. To tackle this challenge, we introduce Knowledge Adapted (\methodname{}) fine-tuning, a data-centric approach that automatically adapts training data with the model's existing knowledge and visual understanding. \methodname{} minimizes hallucinations while preserving high descriptiveness. We validate this approach across several small-scale VLMs (up to 7B parameters) and dense caption datasets, demonstrating that \methodname{} effectively balances hallucination reduction and descriptiveness. Our results show that \methodname{} outperforms various baselines in both automatic metrics and human evaluations.
The code is available \href{https://github.com/moranyanuka/KnowAda}{here}.

\end{abstract} 

\begin{figure}[hbt!]
    \centering
    \includegraphics[trim={7.7cm 2.1cm 4.9cm 1.7cm}, clip, width=1\linewidth]{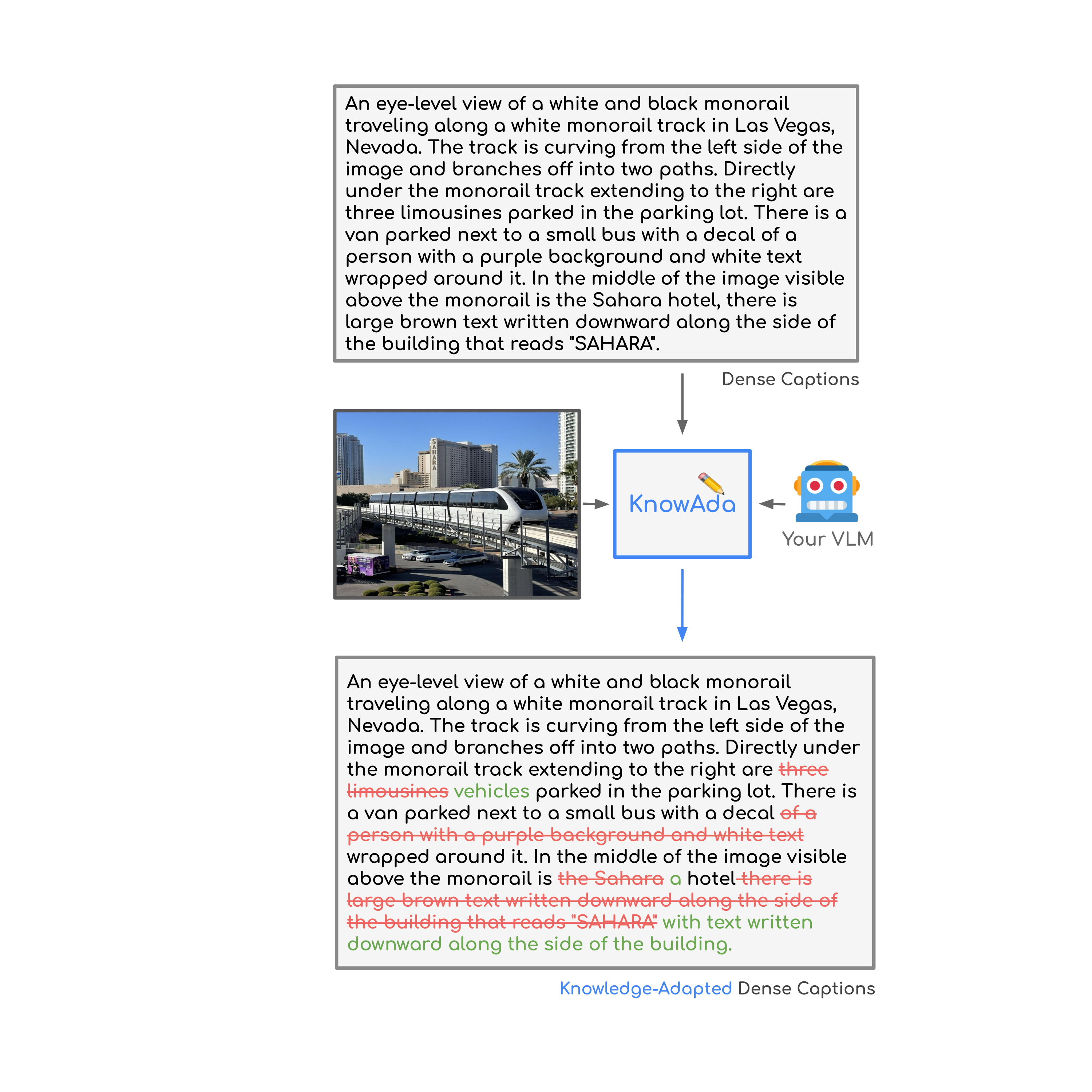}
    \caption{\methodname{} identifies knowledge gaps of a VLM and adapts the dense caption accordingly. The \methodname{} dense captions are better suited for downstream fine-tuning of the VLM.}
    \label{fig:teaser}
\end{figure}

\begin{figure*}[t]
    \centering
    \includegraphics[trim={0.5cm 1cm 0.3cm 1.7cm},clip,width=1.0\linewidth]{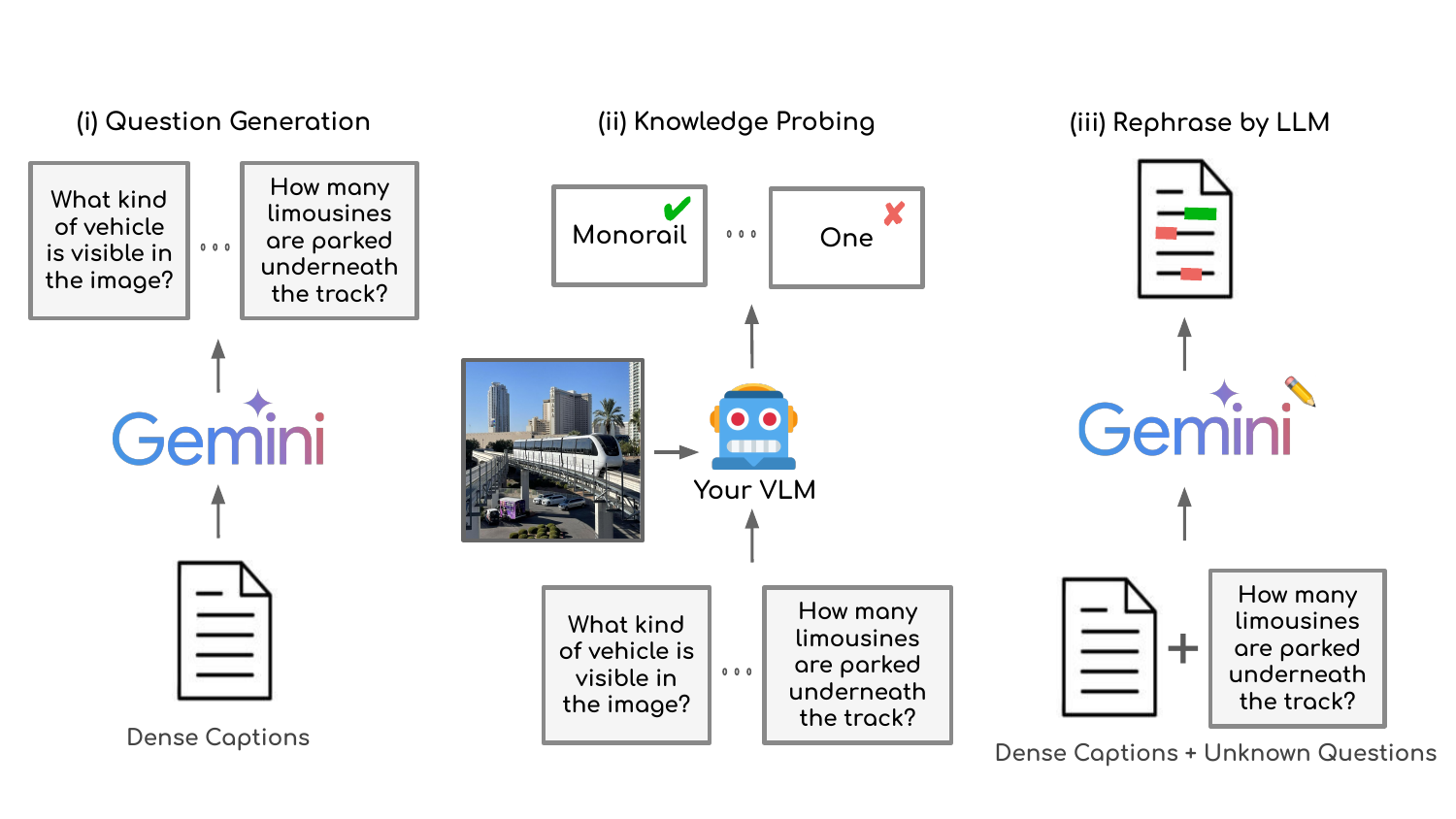}
    \caption{\textbf{Our proposed \methodname{} pipeline.} We first probe the knowledge of the VLM, identifying the known and unknown parts of the image description, by generating questions about the visual content of the image mentioned in the caption. Then, \methodname{} identifies the knowledge gaps by judging the VLM answers to these questions. Finally, \methodname{} adapt the description to match these gaps (e.g., removing the number of limousines mentioned in the caption, which relates to a question the model failed to answer).} 
    \label{fig:pipeline}
\end{figure*}

\section{Introduction}

Fine-tuning pretrained multimodal models for generating dense image captions is common in both research and practical applications, such as assisting visually impaired individuals. Recent work has focused on creating high-quality, descriptive captions through human annotations~\cite{onoe2024docci, garg2024imageinwords, deitke2024molmo} and synthetic generations from models like GPT-4~\cite{chen2023sharegpt4v, chen2024allava} and Gemini~\cite{singla2024pixels}, including extensions that integrate expert models for enhanced detail~\cite{li2024densefusion}. These datasets enable the fine-tuning of models to create detailed descriptions in specific styles, distinguishing them from the zero-shot capabilities of pretrained models. However, smaller multimodal models (e.g., up to 7 billion parameters), which are essential for real-time applications, frequently face challenges in capturing fine-grained visual details during fine-tuning, resulting in hallucinations.

Consider a fine-tuning dataset for dense captioning, like the one in Figure \ref{fig:teaser} from DOCCI~\cite{onoe2024docci}, alongside a pretrained vision-language model trained on tasks like captioning, VQA, and OCR. If, for instance, the model has only encountered low-resolution images during pretraining, it may struggle to identify fine details, such as the drawing on the purple van or the hotel name in the background, which require higher resolution. This issue extends beyond resolution to other visual challenges in modern VLMs~\cite{tong2024eyes, wu2024vsp, zhang2024exploring}. We hypothesize that fine-tuning the model on overly complex captions may increase hallucinations, as the model is compelled to predict details it cannot accurately perceive or understand.

Recent works on large language models (LLMs) have shown that fine-tuning primarily adapts pre-existing factual knowledge for specific tasks, with the majority of this knowledge being encoded during the pretraining phase~\cite{geva2020transformer, meng2022locating, roberts2020much}. Rather than acquiring new information, fine-tuning typically \emph{activates and refines pretrained knowledge}, which remains largely stable throughout the process~\cite{zhou2024lima}. Furthermore, \citet{gekhman2024does} demonstrated that fine-tuning on content not grounded in a model's pre-existing factual-knowledge can lead to an increase in hallucinations. Similarly, \citet{yu2024rlhf} showed that attempting to distill GPT4V into a smaller, less capable VLM significantly increases hallucinations. Motivated by these findings, we hypothesize that one possible cause of increased hallucinations is the excessive visual complexity of captions relative to the model's pretrained capabilities. We propose a method to mitigate this effect.

To better adapt pretrained models to dense caption datasets, we introduce \methodname{}, a model-specific adaptation technique that simplifies complex details in dense captions. The \methodname{} pipeline, shown in Figure \ref{fig:pipeline}, automatically identifies visual knowledge gaps between a pretrained VLM and an image-caption pair by generating questions related to the image's visual content. It then modifies the captions to align with the model's visual knowledge and capabilities, producing adapted captions that enhance fine-grained control while balancing a low hallucination rate with high descriptiveness.

Evaluating dense image captions requires attention to two critical factors: descriptiveness, which captures the image's details, and hallucination rate, which measures factual accuracy. Traditional metrics, such as those based on similarity to reference captions~\cite{papineni2002bleu,banerjee2005meteor,zhang2019bertscore,reimers2019sentence} or CLIP-based alignment~\cite{sarto2023positive,radford2021learning}, often fall short when applied to long captions. These approaches penalize valid variations in phrasing and fail to distinguish between factual accuracy and token overlap or semantic similarity, rendering them inadequate for identifying hallucinations in detailed captions. Existing hallucination metrics are similarly limited, focusing primarily on short captions or object-level errors~\cite{rohrbach2018object,li2023evaluating,ben-kish-etal-2024-mitigating}, or relying on other VLMs~\cite{jing2023faithscore,liu2023mitigating}, which themselves can hallucinate during the verification process. To address these limitations, we propose Decomposed NLI (DNLI), a novel evaluation framework that breaks captions into propositions and assesses their entailment with the detailed ground truth description. DNLI offers a more reliable measure of both descriptiveness and accuracy, demonstrating strong alignment with human judgments.

Our results show that training with \methodname{} captions offers a favorable balance between descriptiveness and fidelity when fine-tuning LLaVA-7B, outperforming other data-centric baselines. To demonstrate the consistency of \methodname{} across different models and datasets, we fine-tuned several multimodal models, ranging from 2 billion to 7 billion parameters, on two different dense captioning datasets. Across all models, \methodname{} consistently reduced hallucinations compared to training with the original captions, as confirmed by both automatic and human evaluations.

Stated explicitly, our contributions are: (I) We show that small-to-medium-scale VLMs underperform when fine-tuned directly on dense caption datasets, exhibiting increased hallucinations and reduced descriptive accuracy; (II) To address this we propose \methodname{}, a model-dependent augmentation method that fills knowledge gaps in captions, reducing hallucinations while preserving high descriptive accuracy;  (III) We demonstrate the effectiveness of \methodname{} through extensive experiments, supported by both quantitative metrics and qualitative human evaluations; (IV) We introduce DNLI, a novel evaluation framework for dense captions that offers a more fine-grained analysis and shows strong correlation with human annotations.

\begin{figure}[t]
    \centering
    \includegraphics[width=1.0\linewidth]{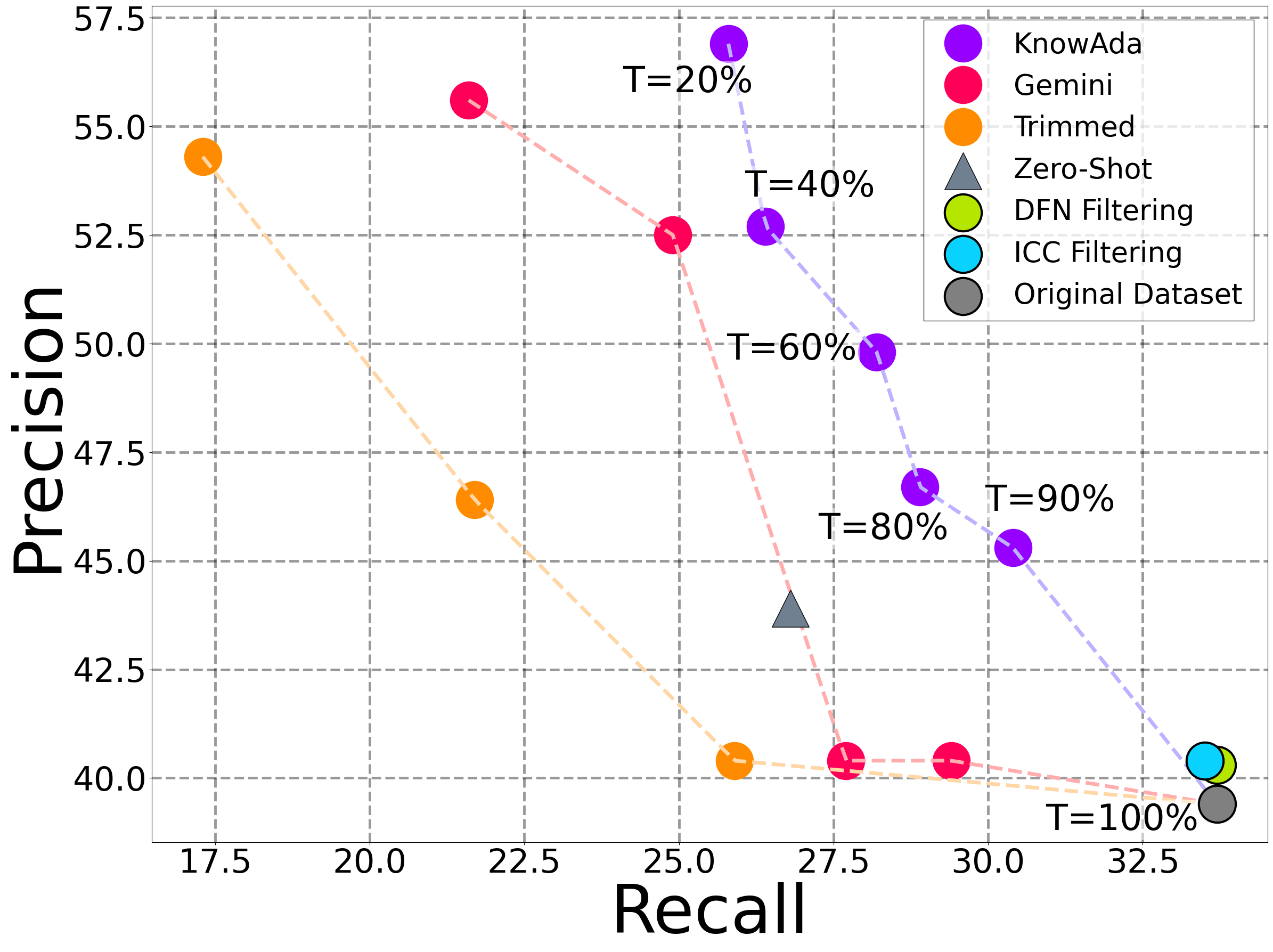}
    \caption{\textbf{Dense captioning descriptiveness precision-recall results} for LLaVA-7B fine-tuned with DOCCI captions, adapted using different methods. ``Trimmed'' refers to naive removal of sentences, while ``Gemini'' involves prompting Gemini to simplify the caption by removing difficult details of varying degrees. \methodname{} consistently achieves better precision-recall balance. Original captions corresponds to \methodname{} with a threshold of $T=100\%$, where no information is classified as unknown.} 
    \label{fig:main_results}
\end{figure}

\begin{figure*}[t]
    \centering
    \includegraphics[width=1.0\textwidth]{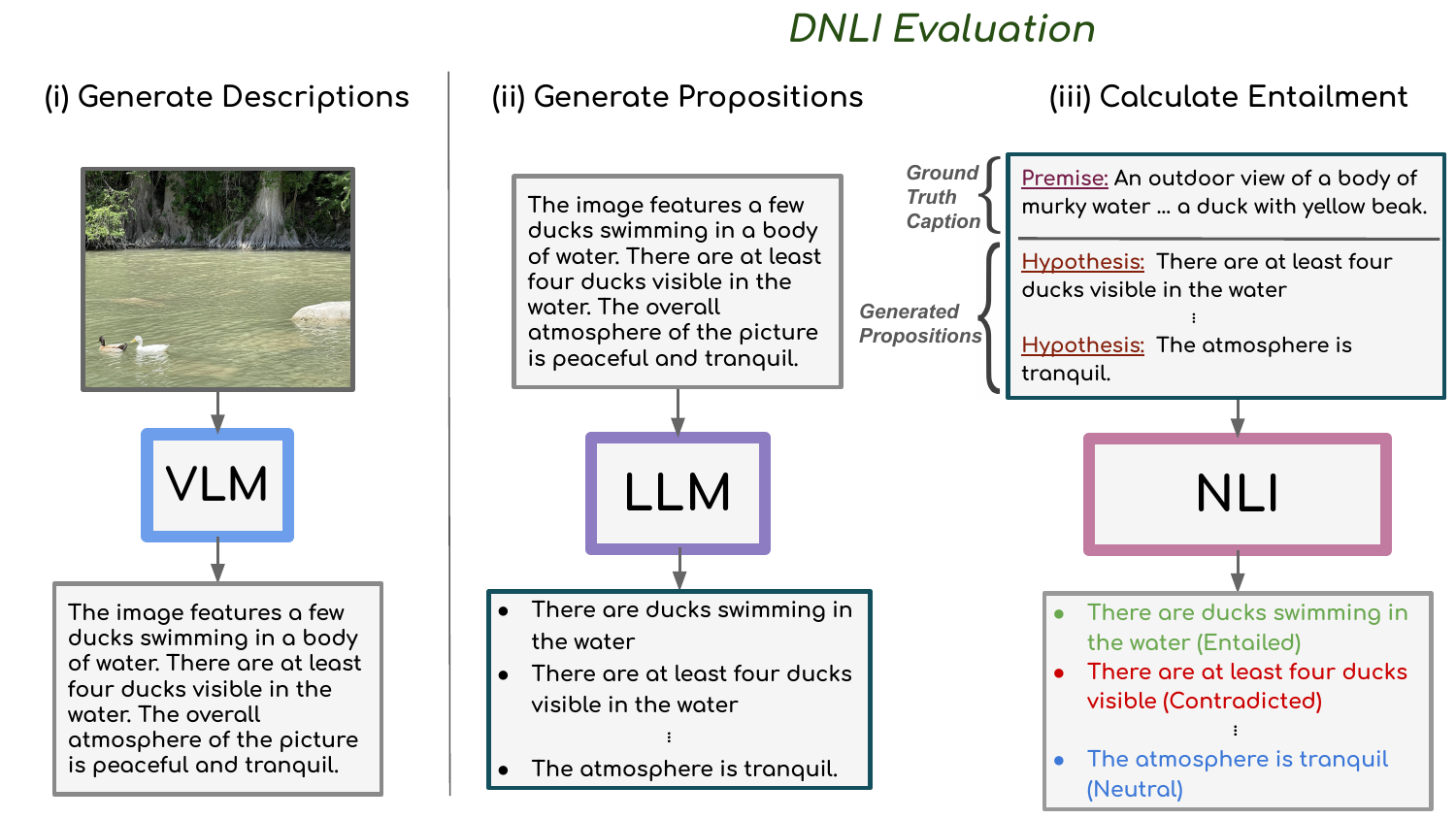}
    \caption{\textbf{DNLI Evaluation}. Given a generated description by a VLM, we decompose it to atomic propositions. Then, we classify each proposition to either entailed, contradicted or neutral, conditioned on the ground-truth description. Finally, we calculate the descriptiveness and contradiction based on the number of entailed and contradicted propositions.}
    \label{fig:eval_pipeline}
    \vspace{-0.1in}
\end{figure*}

\begin{table*}[t]
  \centering
  \setlength{\tabcolsep}{3pt}
  \begin{tabular}{ll|cc|cc|cc|cc|c}
      \toprule
     \multicolumn{1}{c}{} & 
     \multicolumn{1}{c}{} & 
     \multicolumn{4}{|c|}{\textbf{Contradiction $\downarrow$}} & 
     \multicolumn{4}{c|}{\textbf{Descriptiveness $\uparrow$}} & 
      \multicolumn{1}{c}{} \\
     
     \cmidrule(lr){3-6} \cmidrule(lr){7-10}
     \multicolumn{1}{c}{\textbf{Model}} & 
     \multicolumn{1}{c}{\textbf{FT Captions}} & 
     \multicolumn{2}{|c}{\textbf{Precision}} & 
     \multicolumn{2}{c|}{\textbf{Recall}} & 
     \multicolumn{2}{|c}{\textbf{Precision}} & 
     \multicolumn{2}{c|}{\textbf{Recall}} & 
     \multicolumn{1}{c}{\textbf{\#  Words}} \\
     
     \cmidrule(lr){3-4} \cmidrule(lr){5-6} \cmidrule(lr){7-8} \cmidrule(lr){9-10}
     \multicolumn{1}{c}{} & 
     \multicolumn{1}{c}{} & 
     \multicolumn{1}{|c}{Auto} & 
     \multicolumn{1}{c}{Human} & 
     \multicolumn{1}{|c}{Auto} & 
     \multicolumn{1}{c|}{Human} & 
     \multicolumn{1}{|c}{Auto} & 
     \multicolumn{1}{c}{Human} & 
     \multicolumn{1}{|c}{Auto} & 
     \multicolumn{1}{c|}{Human} & 
     \multicolumn{1}{c}{} \\
     \midrule
    PaliGemma  & Synthetic  & 38.9  & 19  & 30 & 15.5  & 47.9  & 81  & \textbf{39.2}  & \textbf{67.8}  & 72 \\
    PaliGemma  & Synthetic KA  & \textbf{32.4}  & \textbf{18.4}  & \textbf{20.8}  & \textbf{13}  & \textbf{55.2}   & \textbf{81.6}  & 36.4  & 61.2  & 54  \\
    
    \midrule
    
    TinyLLaVA  & Synthetic  & 38.1  & 52.1  & 49.1  & 45  & 35.1  & 47.9 & \textbf{26.3}  &  40.8 & 71 \\
    TinyLLaVA  & Synthetic KA  & \textbf{22.9}  & \textbf{34}  & \textbf{40.2}  & \textbf{22.2}  & \textbf{48.1}  & \textbf{66}  & 25.4  & \textbf{40.9}  & 51  \\

    \midrule
    
    LLaVA-7B  & Synthetic & 39.1  & 19.3  & 39.1  & 16.2  & 47.3  & 80.7  & \textbf{39.7}  & \textbf{65.3}  & 79 \\
    LLaVA-7B  & Synthetic KA  & \textbf{31}  & \textbf{15.8}  & \textbf{31}  & \textbf{9.3}  & \textbf{58.4}  & \textbf{84.2}  & 34.5  & 47.4  & 54   \\
    \midrule
    PaliGemma  & Human  & 41.6  & 20.4  & 24.6   & 14.9  & 46.7  & 79.6  & 24.6  & \textbf{43.2}  & 62  \\
    PaliGemma  & Human KA  & \textbf{38.3}  & \textbf{18.3}  & \textbf{22.2}  & \textbf{13.6}  & \textbf{49.4}  & \textbf{81.7}  & \textbf{28.7}  & 38.9  & 65 \\
    \midrule

    TinyLLaVA  & Human  & 53  & 51.8  & 39.5  & 38.8  & 31.9  & 51.8  & \textbf{22.4}  & \textbf{31.4}  & 100  \\
    TinyLLaVA  & Human KA  & \textbf{42.6}  & \textbf{34.5}  & \textbf{19.6}  & \textbf{12.5}  & \textbf{46.9}  & \textbf{65.5}  & 19.1  & 22.5  & 53  \\

    \midrule

    LLaVA-7B  & Human  & 47.2  &  33.4  & 39.7  & 33.2  & 39.4  & 66.6  & \textbf{33.7}  & \textbf{48.1} & 109 \\
    LLaVA-7B  & Human KA  & \textbf{33.7}  & \textbf{17.1}  & \textbf{16.7}  & \textbf{11.2}  & \textbf{56.9}  & \textbf{82.9}  & 25.8  & 31.8  & 55 \\

    \bottomrule
  \end{tabular}
  \caption{\textbf{Dense captioning results} over the test sets of DOCCI when fine-tuning on original human-annotated captions, synthetic captions, and \methodname{}-adapted captions (denoted as KA) with a threshold of 20\%. ``Automatic (Auto)'' refers to model-based NLI evaluation, while ``Human'' refers to evaluations based on human labeling.}
  \label{tab:main_results_new}
\end{table*}

\section{\methodname    }
\label{sec:method}
We begin by introducing our caption adaptation method for dense captioning datasets. \methodname{} comprises three stages, as shown in Figure~\ref{fig:pipeline}. First, we use an LLM to generate visual questions from each dense caption. Next, these questions are employed to probe the VLM's pretrained visual knowledge and identify parts of the image caption the model struggles with. Finally, an LLM adapts the image descriptions by editing out the unknown parts. Below, we elaborate on each step of the pipeline.
\subsection{VLM Knowledge Probing}

In order to detect the visual attributes of the image that are unknown to the VLM but are mentioned in the image description, we probe the VLM knowledge. It is done by generating visual questions which can be answered by the image description, then letting the VLM answer these questions, and finally measuring if the responses are correct. We provide further details regarding each stage below.

\medskip
\noindent\textbf{Finding the unknown questions.} \
Given a dataset \( D \) containing image descriptions, we aim to find all the parts of the description that are unknown to the VLM. Following prior work on uncertainty estimation~\cite{cheng2024can, gekhman2024does}, the steps are as follows:

\vspace{0.05in}
\noindent 1. For each image caption \( d \in D \), generate \( n \) questions \( Q = \{q_1, q_2, \ldots, q_n\} \). Note that $n$ is caption dependent (e.g.,  more questions are generated for longer captions).
\vspace{0.05in}

\noindent 2. For each question \( q_i \in Q \), sample \( m \) answers. Let \( A_i = \{a_{i1}, a_{i2}, \ldots, a_{im}\} \) represent the set of \( m \) answers for question \( q_i \).

\vspace{0.05in}
\noindent 3. Evaluate each answer \( a_{ij} \in A_i \) to determine how difficult it is for the specific VLM. Let \( C_i \subseteq A_i \) be the set of correct answers and \( I_i \subseteq A_i \) be the set of incorrect answers for question \( q_i \).

\vspace{0.05in}
\noindent 4. Calculate the difficulty of each question \( q_i \) based on the ratio of incorrect to incorrect and correct answers. The difficulty score \( df_i \) for question \( q_i \) is given by:
   \[
   df_i = \frac{|I_i|}{|C_i| + |I_i|}
   \]
Here, \( |C_i| \) and \( |I_i| \) are the number of correct and incorrect answers respectively. A higher value of \( df_i \) indicates a more difficult question. For a threshold $T$, a question with $df_i > T$ is defined as an \emph{unknown question}.

We assess the accuracy of the model's responses by prompting a LLM to evaluate each generated answer in relation to the given question and the ground truth description.

For example, consider the two questions shown in Figure \ref{fig:pipeline} (ii). If the model correctly answered \texttt{``Monorail''} for the question \texttt{``What kind of vehicle is visible in the image?''} in 6 out of 10 sampled instances, with 4 incorrect answers, the difficulty for this question is calculated as $df = \frac{4}{10}$ . For the second question, \texttt{``How many limousines are parked underneath the track?''}, the model correctly answered \texttt{``3''} only 4 out of 10 times, resulting in a difficulty of $df = \frac{6}{10}$.
With a threshold of $T=50\%$, the first question is classified as known, while the second as unknown. However, if the threshold is raised to $T=70\%$, both questions would be considered unknown.

For example, consider the two visual questions shown in Figure \ref{fig:pipeline} (ii) with the following prediction accuracies:

\begin{itemize}[itemsep=0pt, parsep=0pt]
    \item For the question \texttt{``What kind of vehicle is visible in the image?''}, the model correctly answers \texttt{``Monorail''} in $6/10$ instances, resulting in $df = 4/10 = 0.4$.
    
    \item For the question \texttt{``How many limousines are parked underneath the track?''}, the model correctly answers \texttt{``three''} in $4/10$ instances, yielding $df = 6/10 = 0.6$.
\end{itemize}

\noindent With a difficulty threshold $T$, the classification of these questions changes:

\begin{enumerate}[itemsep=0pt, parsep=0pt]
    \item For $T = 50\%$:
    \begin{itemize}
        \item The first question is Known (since $df = 0.4 < 0.5$)
        \item  The first question is Unknown (since $df = 0.6 > 0.5$)
    \end{itemize}
    
    \item For $T = 70\%$:
    \begin{itemize}
        \item The first question is Unknown (since $df = 0.4 < 0.7$)
        \item The second question is Unknown (since $df = 0.6 < 0.7$)
    \end{itemize}
\end{enumerate}

\smallskip
\noindent\textbf{Image description rewriting.}
After identifying the questions unknown to the model for each image description, we prompt a LLM, namely, Gemini with several in-context examples to remove or edit the parts of the caption that answer these unknown questions, while keeping the other details intact. For instance, in Figure \ref{fig:teaser}, the questions about the number and type of cars parked in the parking lot is unknown to the small VLM. Therefore the corresponding questions are passed to Gemini, which edits \texttt{``three limousines''} to \texttt{``vehicles''}.

Formally, for each image description \( d_i \), we use all the corresponding questions classified as unknown to the VLM according to the threshold $T$, denoted as \( Q{_i} = \{q_{i,1}, q_{i,2}, \ldots, q_{i,\bar{n}}\} \), to prompt the LLM to rewrite the description by removing information associated with $Q{_i}$. See Appendix \ref{sec:implementation-details} for the prompts used at each stage.

\subsection{\methodname{} Data Analysis}

We start by examining the characteristics of the dense captioning datasets, the impact of \methodname{} on the adapted captions, and the types of questions utilized for visual probing within the \methodname{} pipeline.

\medskip \noindent\textbf{Datasets.}
We use two variations of dense caption datasets in our experiments: DOCCI~\cite{onoe2024docci}, a human-annotated dataset rich in visual details, and DOCCI images paired with synthetic captions generated by Gemini-Pro-1.5, which are designed to be highly visually descriptive. These datasets differ in caption style and level of visual detail, allowing us to demonstrate the robustness of \methodname{} across varying data distributions.

\begin{figure}[t]
    \centering
    \includegraphics[width=1.0\linewidth]{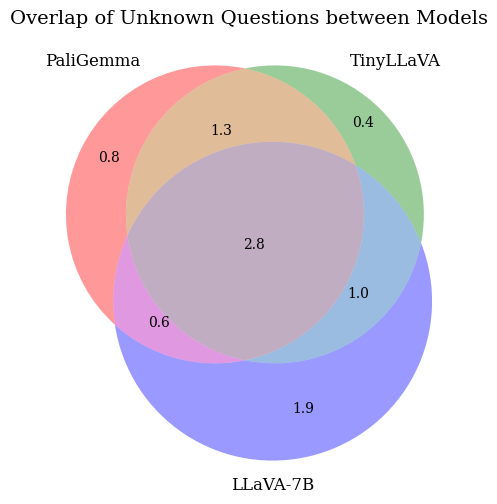}
    \vspace{-0.5pt}
\caption{Overlap in $\bar{Q}_{unk}$ (average number of unknown questions per caption) across different models on the DOCCI training set for $T=20\%$. Each model has a unique set of unknown questions, with about 50\% of these shared among all three models.}
    \label{fig:q_unk_overlap}
\vspace{-0.1in}
\end{figure}

\medskip \noindent\textbf{Dataset characteristics.}
Figure \ref{fig:q_unk_overlap} illustrates the overlap of unknown questions across the different models. While there is a core set of unknown questions common to all three models, each model also has its own unique set of unknown questions.

Table \ref{tab:questions_distribution} presents statistics for each dataset, including the average number of unknown questions per model and dataset, as well as the average word count in both the original and \methodname{} captions. The data indicates that the average number of unknown questions and the average word count in the rewritten captions are relatively consistent across different models, where the human-authored DOCCI captions containing slightly more challenging questions compared to the synthetic captions.

\begin{figure}[t]
    \centering
    \includegraphics[width=1.0\linewidth]{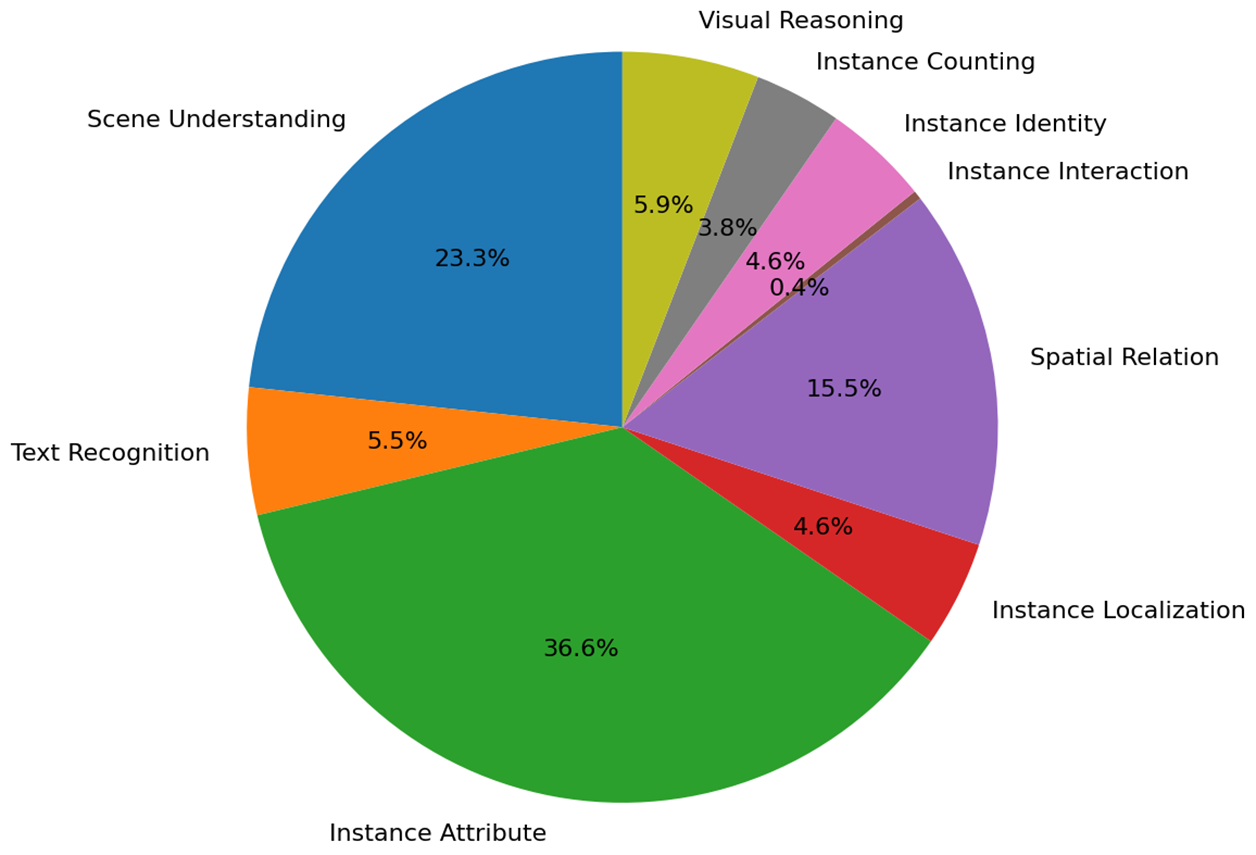}
   \caption{\textbf{Distribution of visual question categories} generated from captions in the first stage of \methodname{}, classified using SeedBench~\cite{li2023seed} definitions.}
    \label{fig:seedbench_question_distribution}
\end{figure}

\medskip \noindent\textbf{Visual questions category distribution.}
To verify that \methodname{} generates diverse types of visual questions, we use LLaMa-3-70B~\cite{dubey2024llama}, quantized to 4 bits, to classify 12,422 questions generated for 1,000 images from the DOCCI test set into categories defined by SeedBench~\cite{li2023seed}. For the classification task, we provide one-few shot examples from each SeedBench category. The resulting distribution of questions is presented in Figure~\ref{fig:seedbench_question_distribution}. As observed, the diversity in the question types suggests that \methodname{} performs knowledge probing across a wide range of visual tasks.

\section{Decomposed NLI Evaluation}
\label{sec:eval_methods}

Next, we introduce DNLI, a proposition-decomposition evaluation framework for dense-captioning. It evaluates the quality of a generated dense image caption using two criteria: \emph{descriptiveness}, which measures accuracy and detail, and \emph{contradiction}, which measures how much the caption contradicts the ground truth. Inspired by prior work on paragraph summarization evaluation~\cite{ernst2021proposition, zhang2021finding, ernst2021proposition} and retrieval~\cite{chen2023dense}, DNLI assesses descriptiveness and consistency through proposition extraction, as outlined below and shown in Figure~\ref{fig:eval_pipeline}. Additional qualitative examples are in the appendix.

\medskip
\noindent
\textbf{Propositional decomposition.}
Given a generated image description, we use a Gemini to decompose it into a set of atomic propositions, capturing individual, verifiable claims. This enables fine-grained evaluation. To avoid duplicates, we instruct the model to include only unique propositions.

\medskip
\noindent \textbf{Natural Language Inference (NLI) analysis.}
We next evaluate the entailment of propositions generated from the image using both our automatic method, where Gemini is used to compute textual entailment, and through human annotators who directly assess visual entailment:

\medskip
\noindent \textit{Automatic Textual Entailment:} \ 
Each atomic proposition is assessed with a Natural Language Inference (NLI) model. This model compares each proposition (the hypothesis) to a ground truth description of the image (the premise), determining whether the proposition is entailed, contradicted, or neutral with respect to the description. This evaluation forms the basis for our descriptiveness and contradiction metrics. See details and prompts in Appendix \ref{sec:implementation-details}.

\medskip
\noindent \textit{Human-Based Visual Entailment Evaluation:} \ We employ Amazon Mechanical Turk to engage three independent annotators for each proposition, tasking them with directly assessing its entailment based on the corresponding image. To ensure quality, we administer a qualification test and select the top-performing annotators. Each proposition is evaluated by a distinct set of three annotators, and the final entailment label is determined by a majority vote. For consistency, we randomly sample 20 images per model, generating an average of 175 propositions per model. In total, 2,812 unique propositions were evaluated. Further details are provided in Appendix~\ref{apx:human_annotation}.

\medskip
\noindent \textbf{Descriptiveness metric.}\ We quantify the \textit{descriptiveness} of the generated caption using two measures: recall and precision.

\smallskip
\noindent  \textit{Descriptiveness Recall:} The proportion of ground truth propositions that are entailed by the generated description:
\[
\text{Descriptiveness Recall} = \frac{|\text{Entailed}|}{|\text{Ground Truth}|},
\]
where the entailed propositions are those identified by the NLI model as described above.

\smallskip
\noindent \textit{Descriptiveness Precision:} The proportion of entailed propositions relative to the total number of propositions in the generated description:
\[
\text{Descriptiveness Precision} = \frac{|\text{Entailed}|}{|\text{Generated}|},
\]
where the generated propositions are those found in the model's caption. The precision represents the likelihood that a given proposition extracted from the generated caption, would be entailed.

\medskip
\noindent \textbf{Contradiction metric.}\ 
The \textit{contradiction} precision and recall are calculated in a similar manner:

\[
\text{Contradiction Precision} = \frac{|\text{Contradicted}|}{|\text{Ground Truth}|}
\]
\[
\text{Contradiction Recall} = \frac{|\text{Contradicted}|}{|\text{Generated}|}
\]

\noindent
Together, these metrics provide a comprehensive evaluation of the generated captions. Descriptiveness precision and recall assess the accuracy and coverage of the content, respectively. Contradiction precision indicates the likelihood of a false proposition in a caption, while contradiction recall measures the total number of contradictions. 

Note that the neutral labels are discarded in the automatic evaluation, as they could represent either subjective propositions (e.g., "vibrant atmosphere") or visual claims not described in the original captions. These labels could be either entailed or contradicted, but it is unclear which.

\begin{table}[t]
\centering
\setlength{\tabcolsep}{4pt}

\begin{tabular}{lcccc}
\toprule
 Captions & Model & \textbf{$\bar{C}_{o}$}  & \textbf{$\bar{C}_{r}$} & $\bar{Q}_{unk}$  \\
\midrule
 Human &  PaliGemma & 122 & 84.2 & 5.5 \\
 Human &  TinyLLaVA & 122 & 80.9 & 6.3 \\
 Human &  LLaVA-7B & 122 & 80.7 & 6.3 \\
\midrule
Synthetic & PaliGemma & 92 & 65.8 & 4.4 \\
Synthetic & TinyLLaVA & 92 & 63.3 &  5.2 \\
Synthetic & LLaVA-7B & 92 & 63.7&  5.2 \\
\bottomrule

\end{tabular}

\caption{Statistics of the original and \methodname{} adapted captions with $T=20\%$ over the human annotated and synthetic version of DOCCI captions. $\bar{C}_{o}$ and $\bar{C}_{r}$ denotes the mean number of words in the original caption and caption after applying \methodname{}, and $\bar{Q}_{unk}$ denotes the mean number of questions unknown to the model.}
\label{tab:questions_distribution}
\end{table}

\begin{table*}[t]
  \centering
  \setlength{\tabcolsep}{7pt}
  \begin{tabular}{c|cc|cc|c}
      \toprule
      \multicolumn{1}{c}{\textbf{Method}} & 
      \multicolumn{2}{c}{\textbf{Contradiction $\downarrow$}} & 
      \multicolumn{2}{c}{\textbf{Descriptiveness $\uparrow$}} & 
      \multicolumn{1}{c}{\textbf{\# Words}}  \\
      \cmidrule(lr){2-3} \cmidrule(lr){4-5}
      \multicolumn{1}{c}{} & 
      \multicolumn{1}{c}{Precision} & 
      \multicolumn{1}{c}{Recall} & 
      \multicolumn{1}{c}{Precision} & 
      \multicolumn{1}{c}{Recall} \\
      \midrule
 \methodname{} Random &  34.7 & 17.2  & 57.9 & 23.9  & 55 \\
 \emph{\methodname{}} & \textbf{33.7}  & \textbf{16.7}  & \textbf{58.9} & \textbf{25.8}  &55 \\
    
    \bottomrule
  \end{tabular}
  \caption{\textbf{Design choices ablations}.We ablate the effect of removing unknown information versus removing random information in the \methodname{} pipeline. Removing unknown information improves performance across all metrics.}
  \label{tab:additional_baselines}
  \vspace{-0.1in}
\end{table*}

\section{Experimental Settings}

Our experiments concentrate on training dense captioning models. We evaluate the performance of models fine-tuned on captions curated through various methods against those fine-tuned on captions adapted using \methodname{}.

We start by comparing \methodname{} to data curation methods for multimodal datasets according to the DataComp challenge~\cite{gadre2024datacomp}, specifically ICC~\cite{yanuka-etal-2024-icc} and DFN~\cite{fang2023data}. 
We also compare our approach to the following baselines that might address the hallucination-descriptiveness trade-off:
\begin{itemize}
\item \emph{Caption Trimming}: A progressive method that removes varying numbers of sentences to simplify captions.
\item \emph{Gemini Simplification}: An approach prompting Gemini to remove difficult details from captions to varying degrees.
\end{itemize}

For all methods in this experiment, we fine-tuned LLaVA-1.5-7B~\cite{liu2024improved} using Low Rank Adaptation~\cite{hu2021lora}.

To ensure that \methodname{} is robust across multiple models and datasets, we fix the threshold at 20\% for classifying questions as unknown and fine-tune three models: PaliGemma~\cite{beyer2024paligemmaversatile3bvlm}, TinyLLaVA~\cite{zhou2024tinyllavaframeworksmallscalelarge}, and LLaVA-1.5-7B~\cite{liu2024improved}. We fine-tune on two DOCCI variations: one using the original DOCCI captions, and another using synthetically generated captions created by Gemini, which were prompted to be visually descriptive. We refer to Appendix \ref{sec:more_experiments} for an experiment on the larger-scale PixelProse~\cite{singla2024pixels} dataset. We evaluate the models using an automatic NLI model and human annotators (Section \ref{sec:eval_methods}). In all experiments, we split the DOCCI test set into 1,000 samples for evaluation, while 4,000 samples are used for the reported test set.

\section{Results}
This section presents our experiments, highlighting the impact of \methodname{} over competing baselines.

\subsection{\methodname{} Achieves Better Descriptiveness-Hallucination Trade-off}
Figure \ref{fig:main_results} illustrates that fine-tuning on \methodname{} captions consistently provides the best balance between high descriptiveness and low hallucination rates compared to competing baselines. While ICC and DFN offer slight improvements in precision, they do not facilitate further gains due to a lack of control over the trade-off. Although trimming and rephrasing captions allow for some control over precision and recall, they yield inferior results compared to \methodname{} across all thresholds.

\subsection{\methodname{} Is Consistent Across Multiple Models and Datasets}
Table \ref{tab:main_results_new} shows that fine-tuning on \methodname{} captions consistently reduces the hallucination rate while maintaining high descriptiveness across different trained models. This superior performance is evident for both human-annotated and synthetically generated dense captions, as confirmed by our automatic and human evaluation pipelines.

Specifically, \methodname{} significantly reduces the contradiction rate in terms of both precision and recall. However, while it improves precision in descriptiveness, it decreases recall. The results in Table \ref{tab:main_results_new} are obtained using a stringent threshold of $T=20\%$, prioritizing hallucination reduction. \methodname{} allows increased recall by enabling the selection of a less stringent threshold, facilitating a trade-off between precision and recall, as illustrated in Figure \ref{fig:main_results}. Future work may explore how to maintain \methodname{}'s improvement in hallucinations while ensuring high recall in descriptiveness.

\subsection{Textual Entailment is Highly Correlated to Human Annotated Visual Entailment}

We computed the Phi correlation coefficient between the majority agreement labels (ground-truth human annotations) and the labels generated by the NLI model for each proposition. The results yielded a Phi coefficient of $\phi=0.73$ for the DOCCI original captions and $\phi=0.67$ for the synthetic captions. These strong positive correlations indicate a significant relationship between the human annotations and our proposed automatic evaluation. It suggests that our text-only NLI model effectively aligns with human judgments of visual entailment, demonstrating its reliability in distinguishing between contradictory and entailed propositions based solely on dense captions.

Moreover, the correlation difference between original and synthetic captions shows that DOCCI human-authored captions are generally more detailed and reliable compared to the synthetic captions, enhancing the effectiveness of the automatic entailment computation using them.

Note that the results from our human annotation process showed an average majority agreement of 82.6\% for the ``Contradicted'' labels and 89.7\% for the ``Entailed'' labels, indicating a high level of consensus among annotators.

\subsection{Necessity of Removing Unknown Information}

\noindent
We perform an ablation study to assess the importance of removing VLM-unknown information by following the \methodname{} procedure for a fixed $T=20\%$, but instead of removing information linked to unknown questions, we randomly remove information, a method referred to as \emph{\methodname{} Random} in Table \ref{tab:additional_baselines}. As expected, this results in worse performance across all metrics, demonstrating that unknown information is indeed a problematic factor in captions that impacts the hallucinations.

\section{Conclusions}
This work focuses on fine-tuning small-to-medium-scale vision-language models by aligning dense captions with the models' existing knowledge and visual understanding. It aims to address the challenge of balancing rich descriptiveness with factual correctness in multimodal models, especially when models have limited capacity to process complex visual details in dense captions. By probing models with visual questions and adapting captions to exclude unknown information, we reduce hallucinations while maintaining high descriptive accuracy. Our results, supported by both human and automatic evaluations, suggest that this strategy can lower hallucinations compared to training on original captions or competing data curation baselines.

Furthermore, we introduce a proposition-based evaluation framework that provides fine-grained analysis of generated captions, offering deeper insights into the balance between descriptiveness and factuality.  We believe that our findings contribute to improving fine-tuning methods in VLMs from a data-centric approach, particularly in resource-constrained environments where balancing descriptiveness and accuracy is important for real-world applications. We hope that this work will encourage further research into addressing the challenges of dense captions in VLMs.

\section{Related Work}

Our research is connected to the field of dense image captioning, particularly in addressing unknown information and enhancing fine-tuning datasets. Below, we provide an overview of previous works in each of these areas.

 \vspace{-2pt}
\subsection{Dense Image Captioning} 
 \vspace{-2pt}
Traditional image captioning datasets like COCO~\cite{chen2015microsoft} often feature brief captions with limited visual detail. Recently, interest has grown in creating longer and more complex captions. For example, \citealt{onoe2024docci} and \citealp{garg2024imageinwords} developed a dataset of 15K human-annotated, richly descriptive captions. \citet{shabtay2024livexivmultimodallive} used captions of figures of arXiv papers. Additionally, \citealp{chen2023sharegpt4v} and \citealp{singla2024pixels} employed GPT-4 and Gemini to generate detailed captions. We show that these datasets are more effective for fine-tuning when adapted to the specific model being trained.

\vspace{-2pt}
\subsection{Training On Unknown Information}
\vspace{-2pt}
\noindent Several works examine the relationship between unknown information and downstream model performance.
\citet{gekhman2024does} show that fine-tuning LLMs on low-confidence examples encourages hallucinations, and suggest that fine-tuning should not introduce new knowledge, but only teach the model to make use of existing knowlegde. 
\citet{zhang2024selfalignmentfactualitymitigatinghallucinations} suggest optimizing for higher-confidence responses via Direct Preference Optimization (DPO), where the confidence of each response is estimated via self-evaluation.
\citet{piché2024llmslearnselfrestraintiterative} train a utility function that encourages predicting only high-certainty responses. The function is trained over synthetic data, which is collected iteratively, and is composed of fused high-confidence predicted facts. Additionally, \citet{xu2024rejectionimprovesreliabilitytraining} introduce a refusal mechanism which encourages the model to reject questions that do not align with its existing knowledge.   We demonstrate that training a model on questions it is fundamentally lacking the ability to answer, rather than solely on factual knowledge it lacks—particularly in the multimodal domain—results in similar performance degradation. Moreover, our method only rejects the unknown parts of a data sample, while maintaining the information that is useful for training, thus avoiding throwing out useful training samples which only include some unknown parts.

 \vspace{-2pt}
\subsection{Data Curation in Fine-Tuning}
 \vspace{-2pt}
Recent research has focused on refining the fine-tuning process for pretrained models, especially through instruction tuning. \citet{zhou2024lima} demonstrated that fine-tuning on a small, high-quality instruction dataset can yield superior results. \citet{li2023quantity} explored training a language model on a subset of instruction data, using loss discrepancies to inform filtering strategies. However, this approach doesn't extend to image captioning, where no instructions are available. \citet{lin2024rho} proposed reducing noisy token influence by assigning lower weights, while \citet{chen2024your} trained a network to filter out low-loss instruction samples. In contrast, our approach adapts specific parts of captions misaligned with the model’s capabilities, rather than filtering or reweighing the entire caption. Our strategy leverages visual knowledge and emphasizes sequences and conceptual coherence, moving beyond token-level adjustments.

\section{Limitations} 
While \methodname{} demonstrates notable effectiveness, several limitations must be considered.

Firstly, \methodname{} is model-dependent, probing the knowledge of each model in isolation. This means that the data must be tailored for each model independently, leading to increased computational overhead.

Secondly, while we show that \methodname{} achieves better performance in balancing the descriptiveness-hallucination trade-off, it does not fully resolve this issue. Reducing contradictions to the ground truth often decreases descriptiveness recall (but not precision). 

Lastly, our analysis focused specifically on dense caption generation, which is a limited task. Expanding \methodname{} to other tasks, such as visual question answering (VQA), could represent an interesting avenue for future research.

\section*{Ethics Statement}
\label{sec:broader_impact}
This work focuses on measuring and mitigating hallucinations in visual-language models (VLMs). As such, it is expected to increase the reliability of VLMs and the ability to measure their performance, which is important when using them in real-world systems. This is expected to have a positive impact on the use of VLMs in society. However, we recognize that the foundation models used in the \methodname{} construction and evaluation pipeline could propagate biases. We anticipate further research into such biases before relying on our work beyond the research environment. The human annotation study performed in this work received the required IRB approval from our institution's ethics committee.

\section*{Acknowledgments}

This work was partially supported by Google and the TAU Center for Artificial Intelligence and Data Science. The authors thank Morris Alper and Nimrod Shabtay for their valuable feedback and assistance.

\bibliography{anthology,custom}

\clearpage

\appendix

\section{Appendix}
\label{sec:appendix}
In this appendix, we describe the implementation details in Appendix \ref{sec:implementation-details}, qualitative examples in Appendix \ref{sec:qualitative_examples}, additional human annotations details in Appendix \ref{apx:human_annotation} and additional experiments in Appendix \ref{sec:more_experiments}.

\begin{table*}[!ht]
\centering
\setlength{\tabcolsep}{4pt}
\begin{tabular}{lccccc}
\hline
\textbf{Model}         & \textbf{Epochs} & \textbf{BS} & \textbf{LoRA Rank} & \textbf{LR} & \textbf{Checkpoint} \\ \hline
LLaVA-1.5-7B           & 3               & 8                   & 64                         & 0.0001      & \texttt{llava-hf/llava-1.5-7b-hf} \\         
TinyLLaVA              & 3               & 8                   & 64                         & 0.0001     & \texttt{tinyllava/TinyLLaVA-Gemma-SigLIP-2.4B} \\            
PaliGemma              & 15              & 8                   & 128                        & 0.0001             & \texttt{google/paligemma-3b-mix-224} \\ \hline   
\end{tabular}
\caption{Training hyperparameters for the models. BS and LR refers to batch size and learning rate.}
\label{tab:hyperparameters}
\end{table*}

\subsection{Implementation Details}
\label{sec:implementation-details}

\begin{figure*}[ht]
    \centering
    \includegraphics[trim={0cm 0cm 0cm 0.35cm},clip,width=1.0\textwidth]{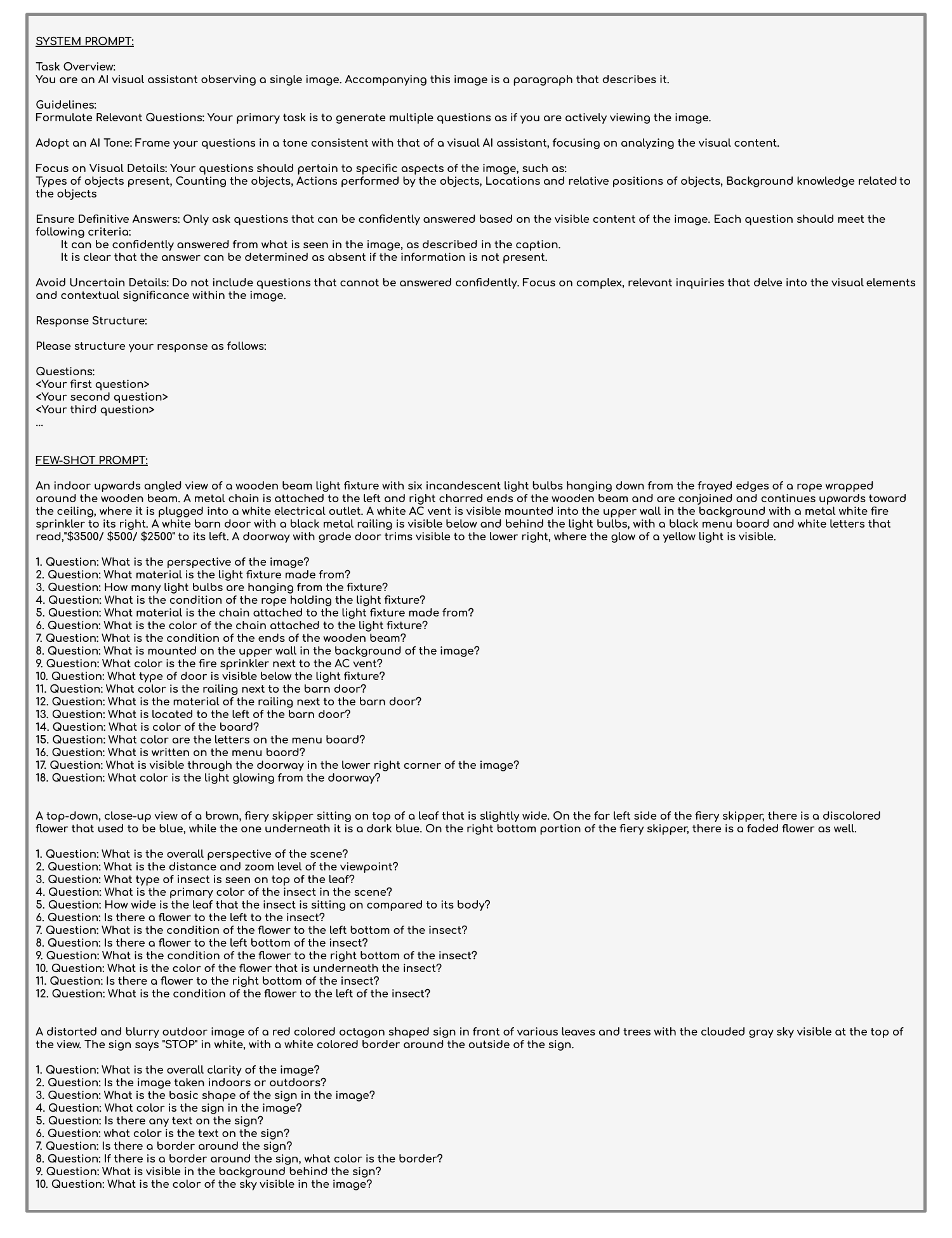}
    \caption{\textbf{Question Generation Prompt}.}
    \label{fig:prompt_quest_gen}
\end{figure*}

\begin{figure*}[ht]
    \centering
    \includegraphics[trim={0cm 0cm 0cm 0cm},clip,width=1.0\textwidth]{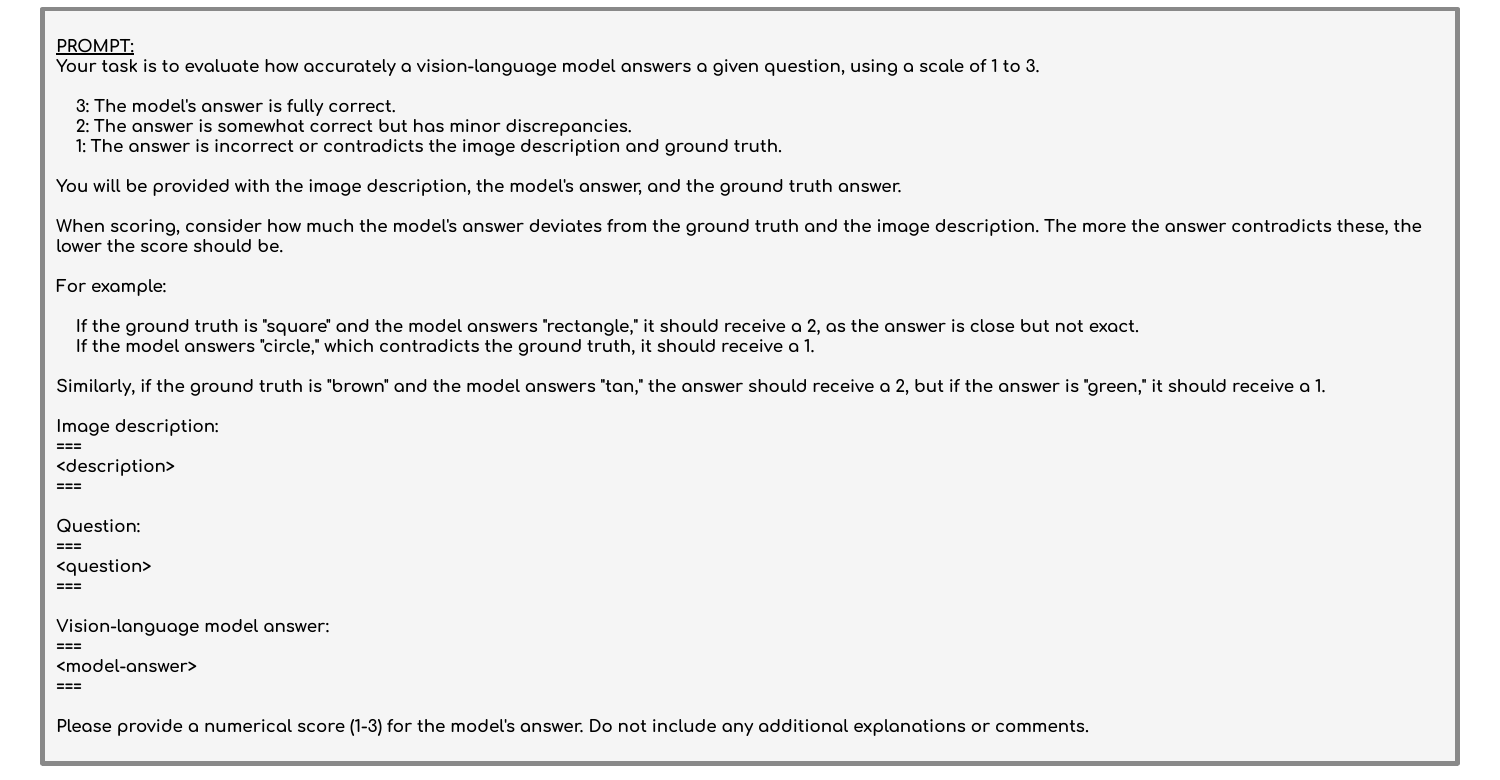}
    \caption{\textbf{Evaluate VLM Answer Prompt}.}
    \label{fig:prompt_evaluate_vlm_ans}
\end{figure*}

\begin{figure*}[ht]
    \centering
    \includegraphics[trim={0cm 0cm 0cm 0.35cm},clip,width=1.0\textwidth]{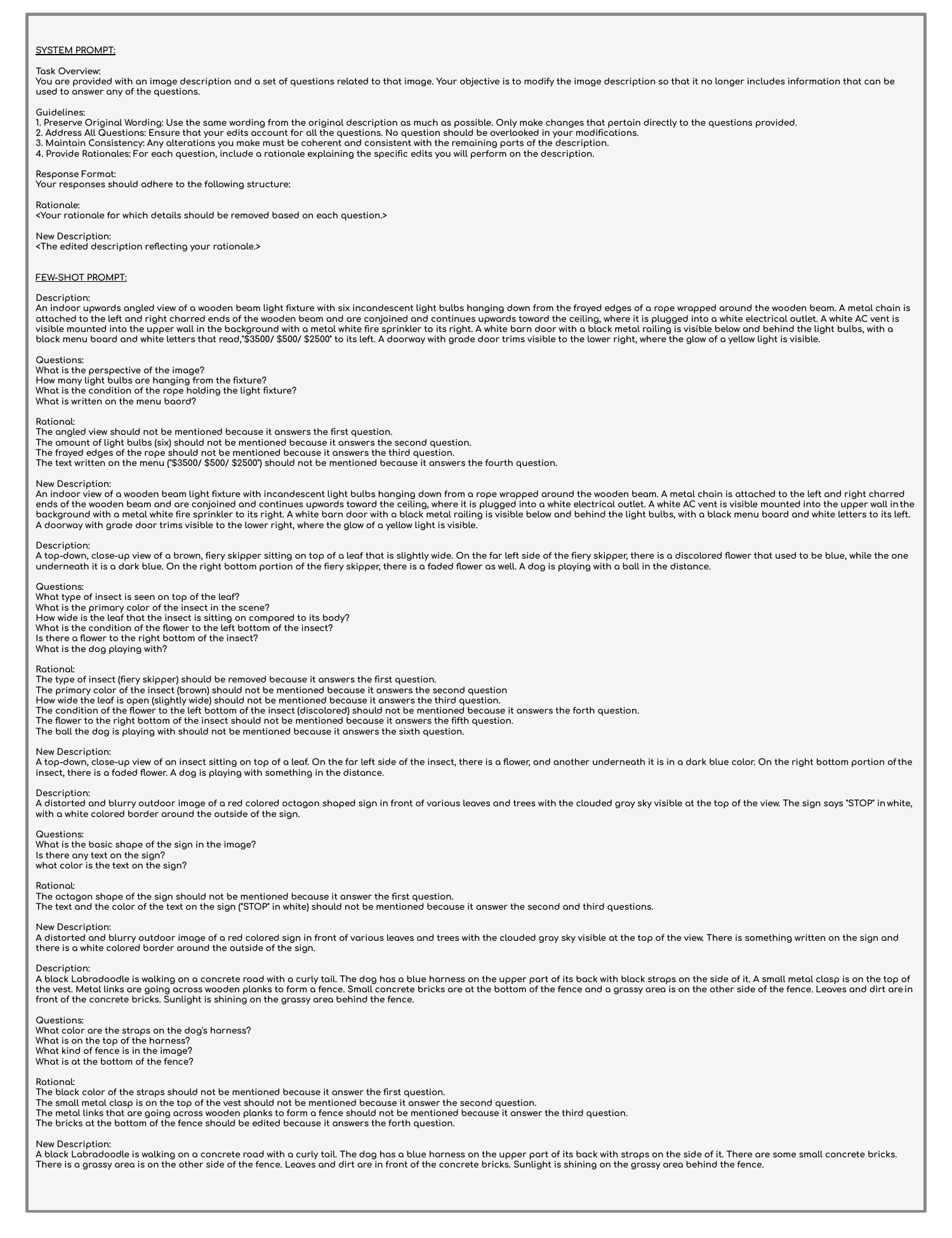}
    \caption{\textbf{Rewriting Prompt}.}
    \label{fig:prompt_rewriting}
\end{figure*}

\smallskip
\noindent
\textbf{\methodname{} Implementation Details} \
For question generation, we use the \texttt{gemini-1.5-flash-001} model, following the prompt in Figure \ref{fig:prompt_quest_gen}. To generate answers, we set the temperature to 0.4 for all models, sampling 10 answers per question. These answers are evaluated using the \texttt{gemini-1.5-flash-001} model, with prompts provided in Figure \ref{fig:prompt_evaluate_vlm_ans}. For rewriting image descriptions, we employ \texttt{gemini-1.5-pro}, utilizing manually curated few-shot examples as illustrated in Figure \ref{fig:prompt_rewriting}. In all experiments, we use greedy sampling when generating outputs with Gemini.

\begin{figure*}[ht]
    \centering
    \includegraphics[trim={0cm 0cm 0cm 0cm},clip,width=1.0\textwidth]{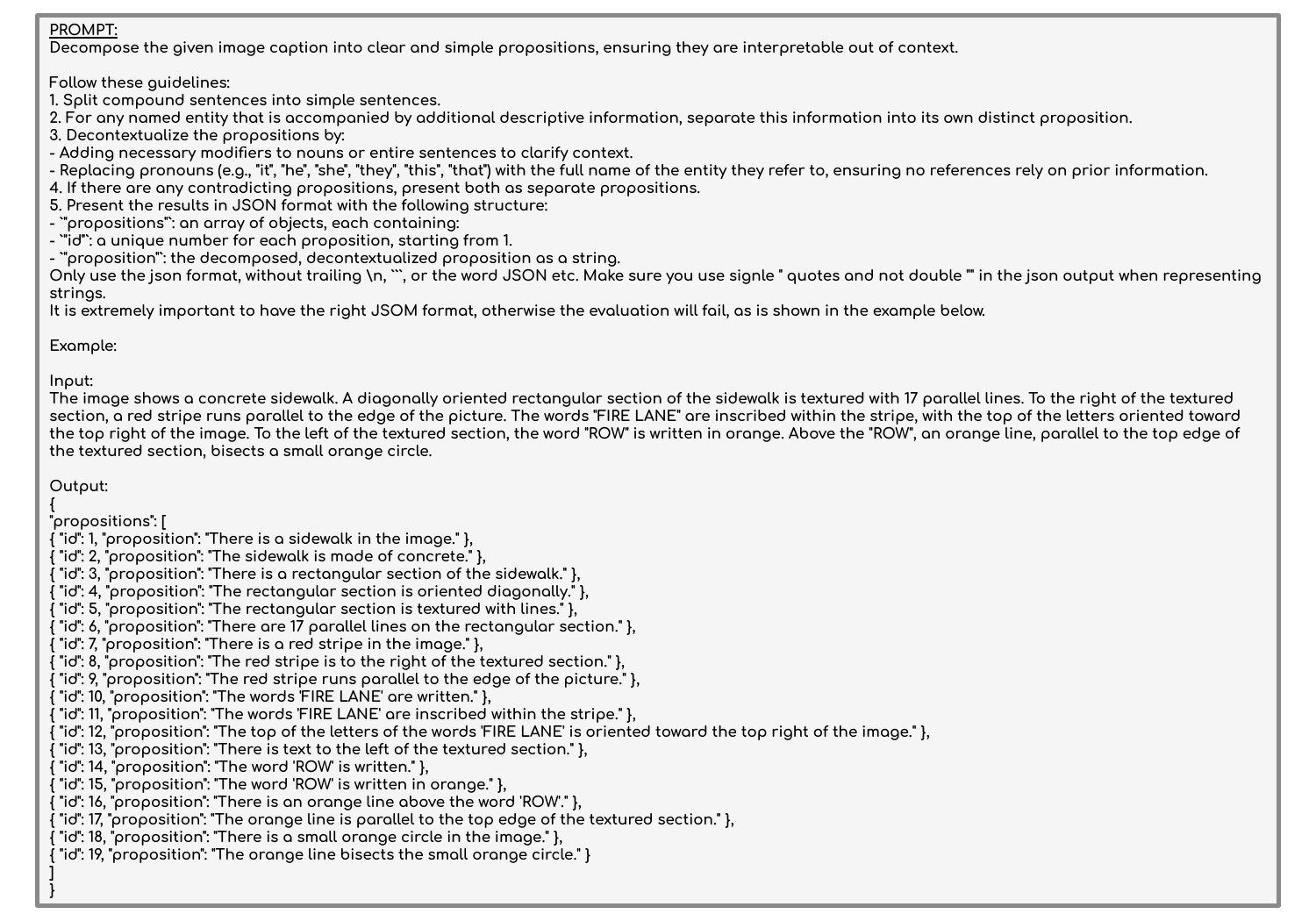}
    \caption{\textbf{Proposition Extraction Prompt}.}
    \label{fig:prompt_proposition}
\end{figure*}

\begin{figure*}[ht]
    \centering
    \includegraphics[trim={0cm 0cm 0cm 0cm},clip,width=1.0\textwidth]{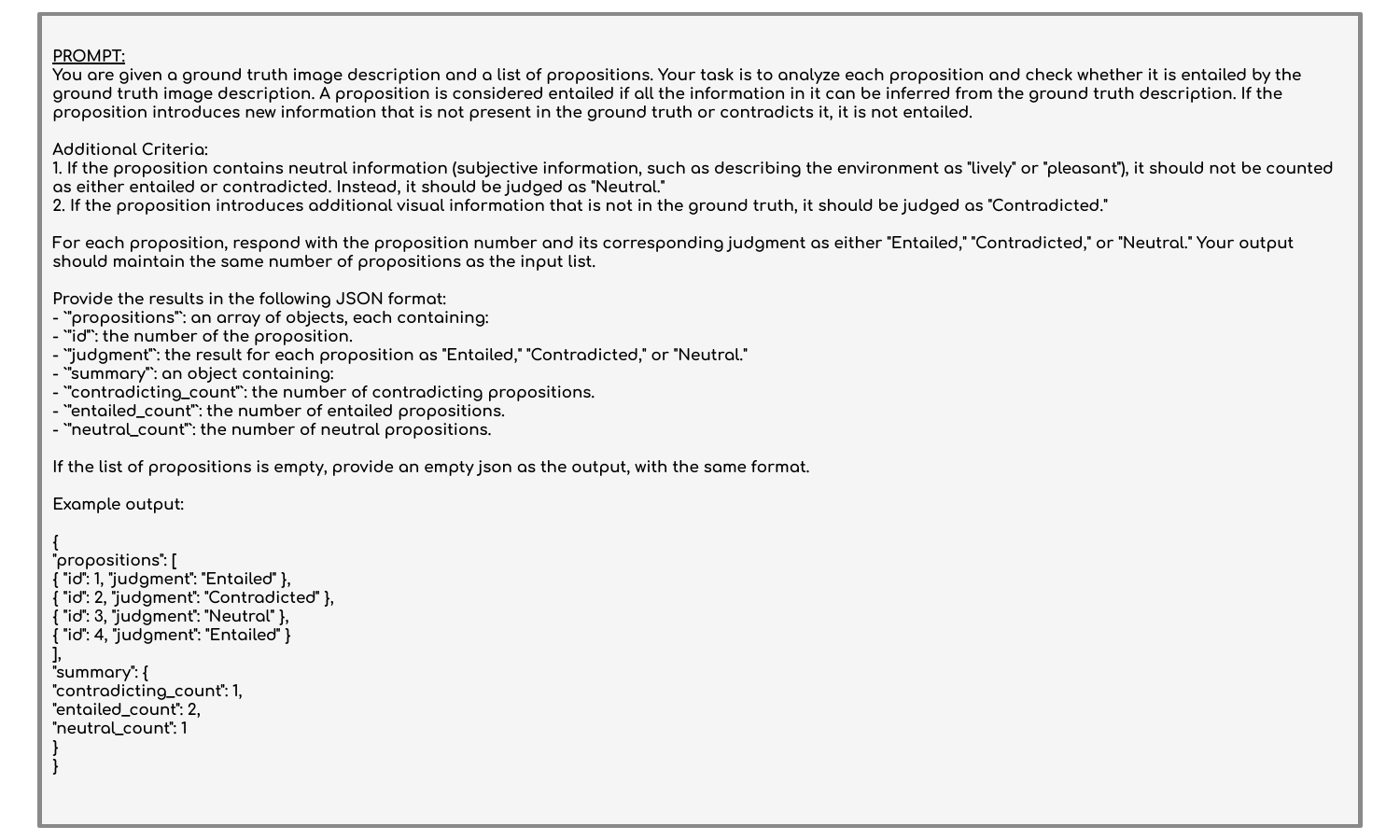}
    \caption{\textbf{Proposition Judgement Prompt}.}
    \label{fig:prompt_judgement}
\end{figure*}

\smallskip
\noindent
\textbf{Proposition Entailment Evaluation Implementation Details} \
We use \texttt{gemini-1.5-flash-001} both for the proposition extraction, as well as the textual entailment task. We constrain the output to be in JSON format, following the proposition extraction prompt shown in Figure \ref{fig:prompt_proposition} and textual entailment prompt in Figure \ref{fig:prompt_judgement}.

\smallskip
\noindent
\textbf{Training Details} \
We provide the training hyperparameters in Table \ref{tab:hyperparameters}, all other hyperparameters are set to default. LLaVA and PaliGemma were trained using LLaMA-Factory framework~\cite{zheng2024llamafactory} and TinyLLaVA was trained using TinyLLaVA-Factory framework~\cite{jia2024tinyllava}.
We note that this work fully complies with the licenses of all used scientific artifacts (e.g. DOCCI, LLaVA, PaliGemma, etc.). All use of scientific artifacts is consistent with their intended use. All models were train on a single NVIDIA A6000 GPU.

\begin{table*}[t]
  \centering
  \setlength{\tabcolsep}{5pt}
  \begin{tabular}{ll|cc|cc|cc|cc|c}
      \toprule
     \multicolumn{1}{c}{} & 
     \multicolumn{1}{c}{} & 
     \multicolumn{2}{|c|}{\textbf{Contradiction $\downarrow$}} & 
     \multicolumn{2}{c|}{\textbf{Descriptiveness $\uparrow$}} & 
     \multicolumn{1}{c}{} \\
     
     \cmidrule(lr){3-4} \cmidrule(lr){5-6}
     \multicolumn{1}{c}{\textbf{Model}} & 
     \multicolumn{1}{c}{\textbf{FT Captions}} & 
     \multicolumn{1}{|c}{\textbf{Precision}} & 
     \multicolumn{1}{c|}{\textbf{Recall}} & 
     \multicolumn{1}{|c}{\textbf{Precision}} & 
     \multicolumn{1}{c|}{\textbf{Recall}} & 
     \multicolumn{1}{c}{\textbf{\# Words}} \\
     
     \midrule
    PaliGemma  & PixelProse  & 44.1 & 40.2  & 43.4 & \textbf{38.4}  & 70.2 \\
    PaliGemma  & PixelProse KA  & \textbf{35} & \textbf{25.4}  & \textbf{54.2} & 38.2  & 47.7 \\
    
    \bottomrule
  \end{tabular}
  \caption{\textbf{Dense captioning results} using the automatic version of DNLI on the PixelProse~\cite{singla2024pixels} test set, when fine-tuning on original captions and \methodname{}-adapted captions (denoted as KA) with a 20\% threshold.}
  \label{tab:pixelprose}
\end{table*}

\begin{figure}[ht]
    \centering
    \includegraphics[width=0.8\linewidth]{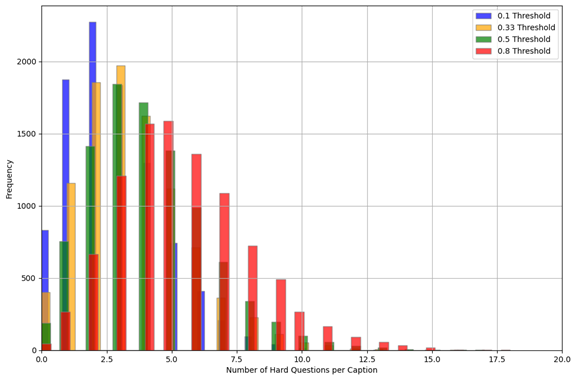}
    \vspace{-10pt}
    \caption{Distribution of number of unknown questions per caption for each ratio threshold of correct and incorrect answers from the vlm.} 
    \label{fig:threshold_questions_distribution}
\end{figure}

\begin{figure}[ht]
    \centering
    \includegraphics[width=0.8\linewidth]{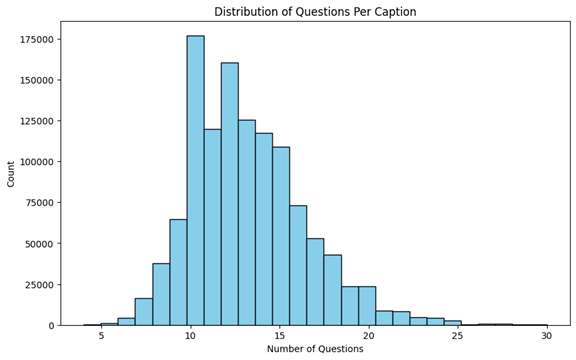}
    \vspace{-10pt}
    \caption{\textbf{Distribution of number of questions per caption} that are generated by the LLM based on the ground truth image description during the first stage of \methodname{}.} 
    \label{fig:questions_distribution}
\end{figure}

\begin{figure*}[ht]
    \centering
    \includegraphics[width=1.0\linewidth]{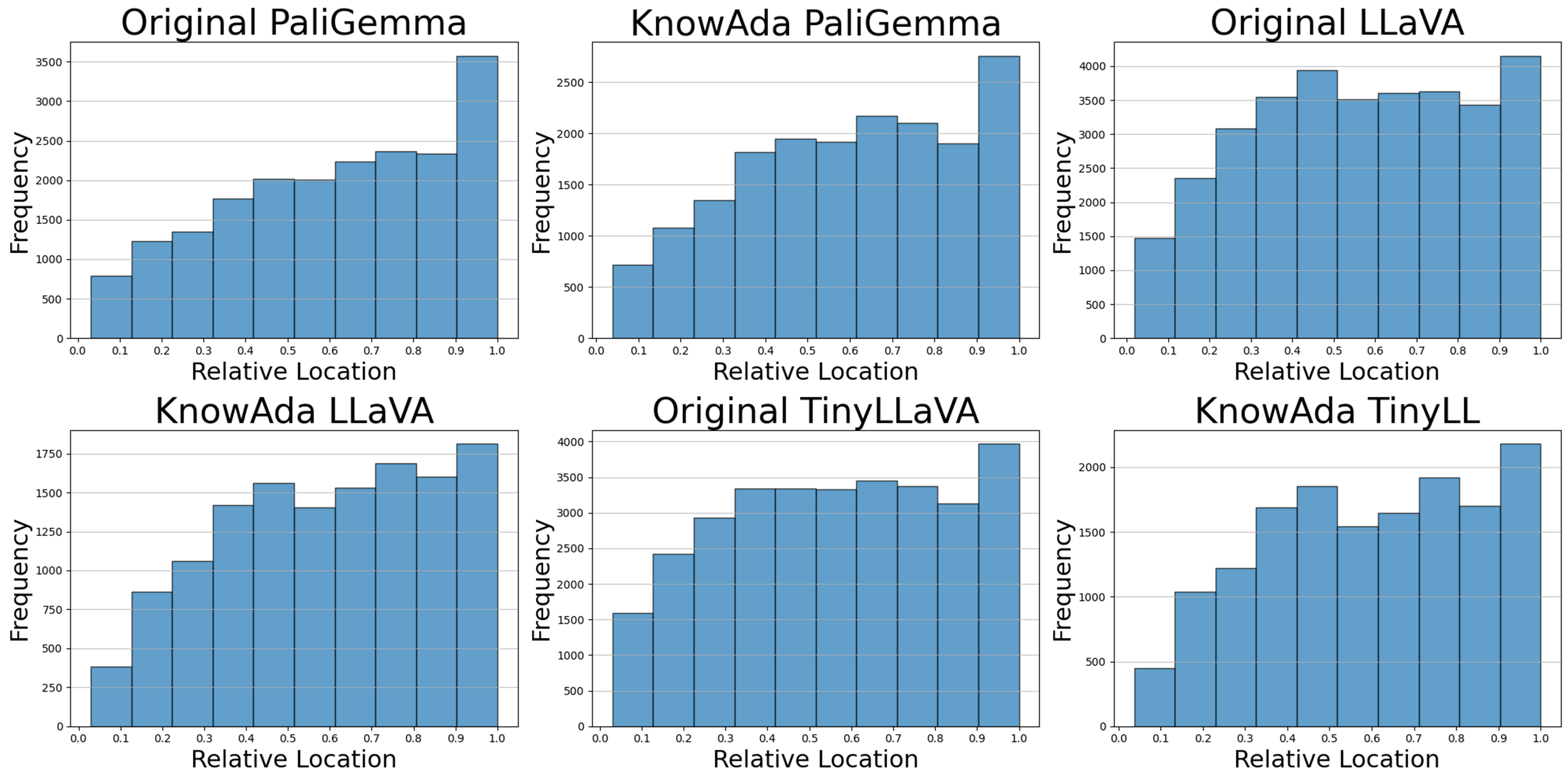}
    \vspace{-10pt}
    \caption{\textbf{Distribution of contradicted propositions' relative location} in generated captions across different models, tested on the DOCCI test set and measured using our proposition evaluation framework. Contradictions increase progressively as the captions become longer.} 
    \label{fig:cont_location}
\end{figure*}

\subsection{Qualitative Examples}
\begin{figure*}[ht]
    \centering
    \includegraphics[trim={0.6cm 1.3cm 0.5cm 1.3cm},clip,width=1.0\textwidth]{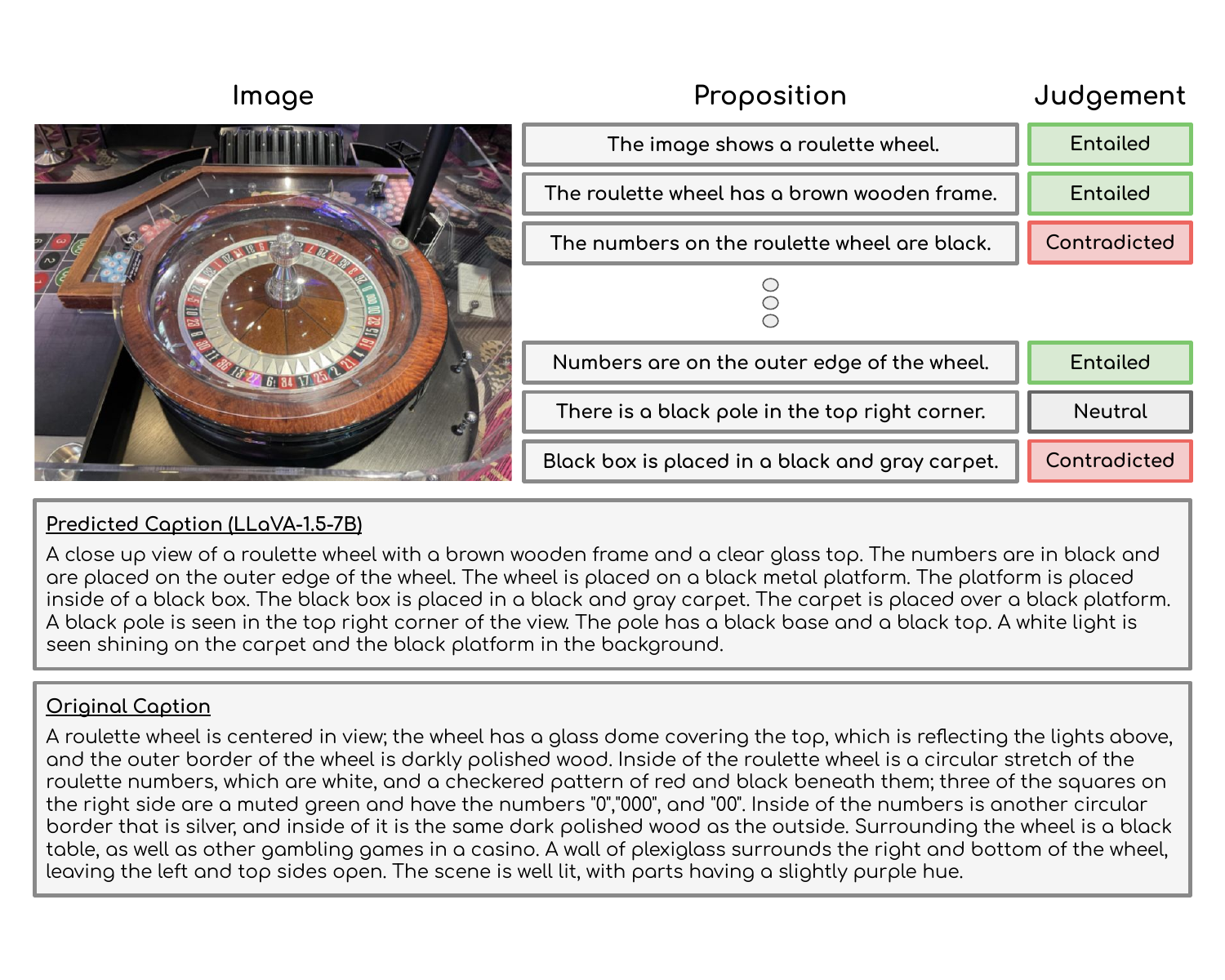}
    \caption{\textbf{Qualitative Example of Our Proposition-Based Evaluation Metric}. Given a generated description by a VLM (middle), we decompose it into atomic propositions (top, center part). Then, we classify each proposition as either 'Entailed', 'Contradicted' or 'Neutral' (top, right part), conditioned on the ground-truth image description (bottom). Finally, we calculate the consistency and contradiction based on the number of entailed and contradicted propositions.}
    \label{fig:props_qual}
\end{figure*}

In Figure \ref{fig:props_qual}, we show example of our evaluation pipeline, including the original description, the generated description and the labels produce by the NLI model.

\begin{figure*}[ht]
    \centering
    \includegraphics[trim={1cm 0cm 1cm 0cm},clip,width=1.0\textwidth]{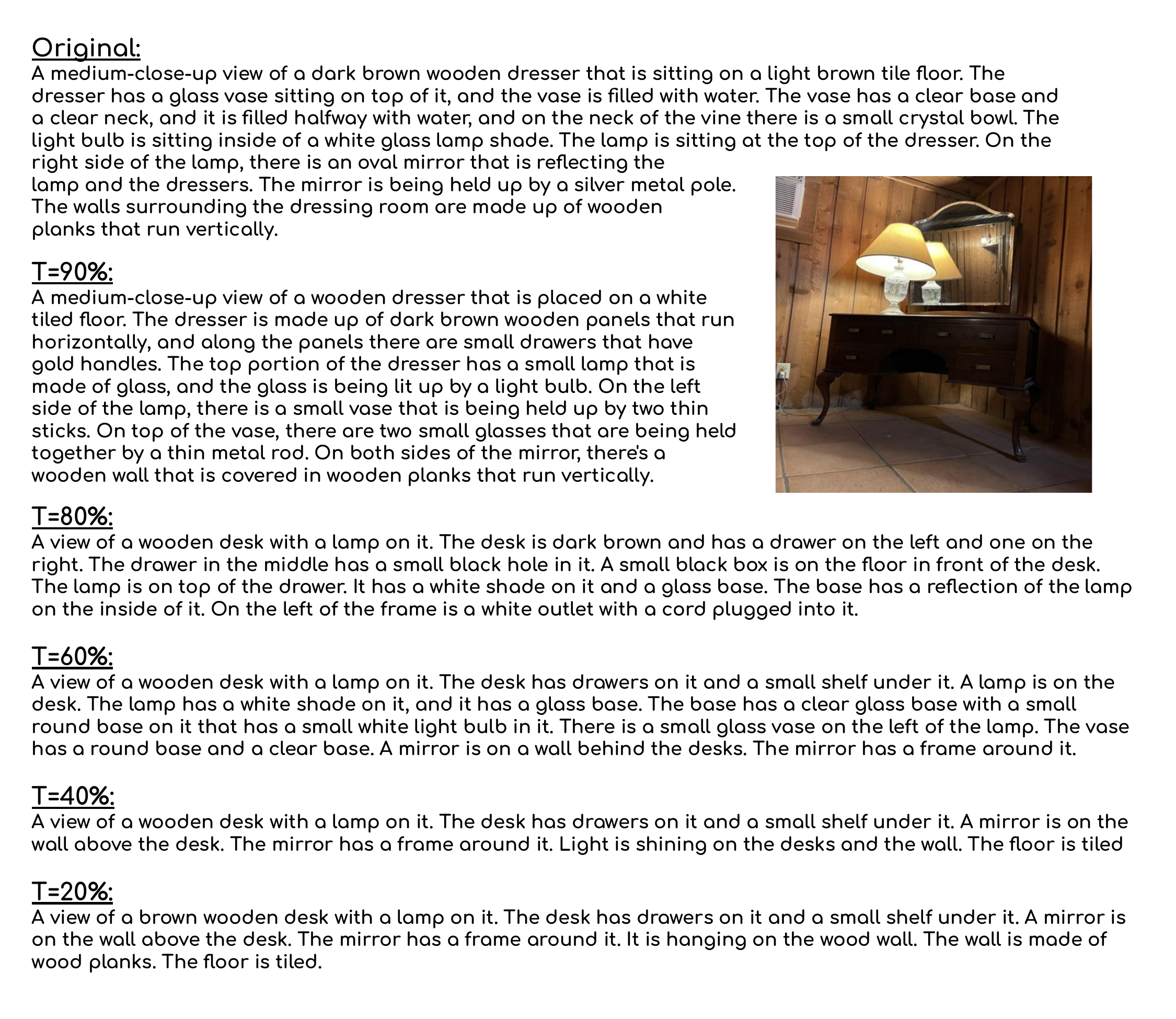}
    \caption{\textbf{Qualitative Example of \methodname{}'s Threshold $T$  effect}. We apply \methodname{}'s rephrasing to the original caption (bottom) using different T values.}
    \label{fig:different_ts_2}
\end{figure*}

\begin{figure*}[ht]
    \centering
    \includegraphics[trim={1cm 0cm 1cm 0cm},clip,width=1.0\textwidth]{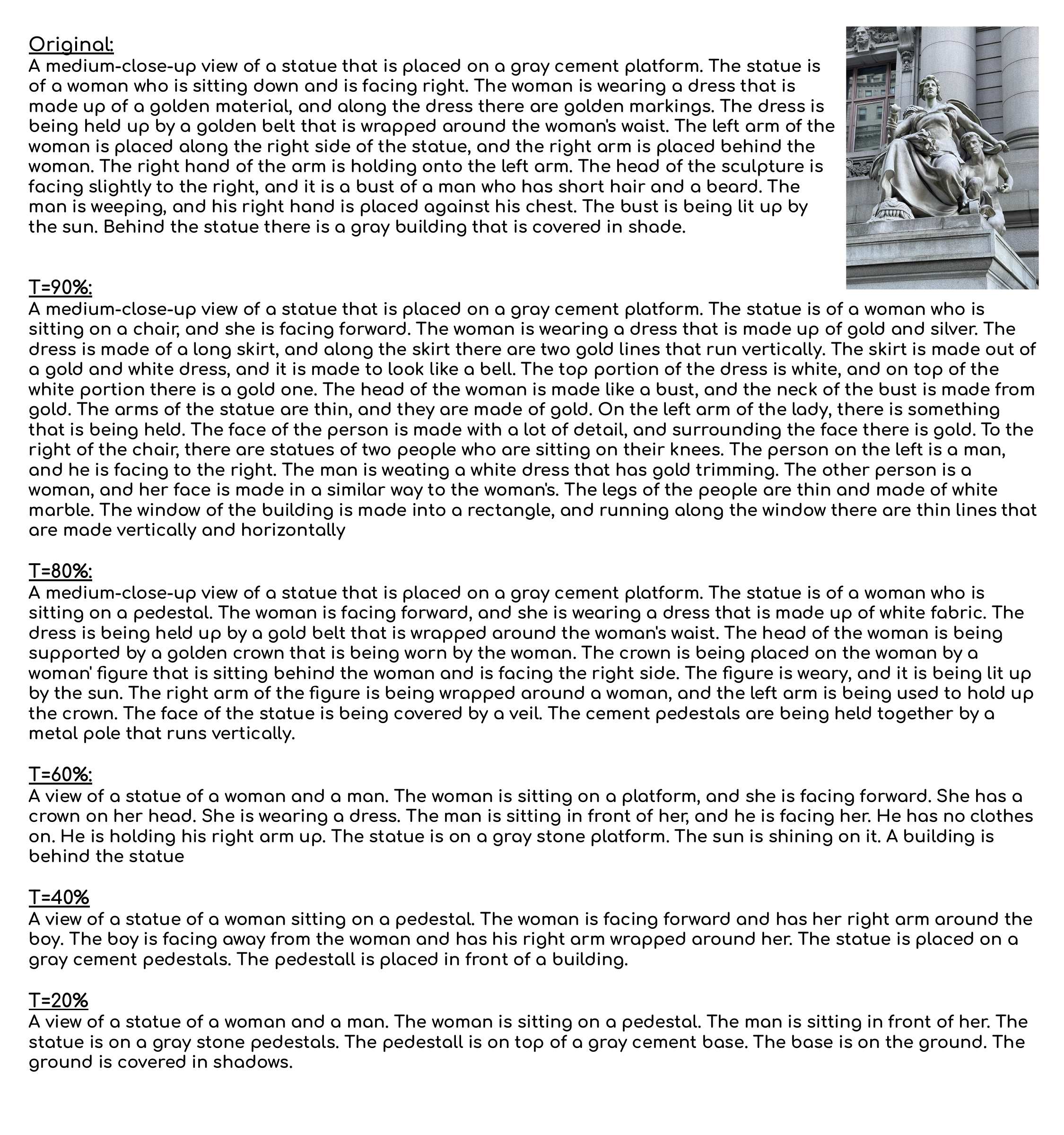}
    \caption{\textbf{Additional Qualitative Example of \methodname{}'s Threshold $T$  effect}.}
    \label{fig:different_ts}
\end{figure*}

\label{sec:qualitative_examples}
Figure \ref{fig:different_ts_2} and Figure\ref{fig:different_ts} shows outputs generated by LLaVA-1.5-7B, trained on \methodname{}-adapted captions using various thresholds. As the threshold $T$ decreases, the model's tendency to hallucinate decreases. However, this comes at the cost of reduced detail coverage. The threshold $T$ allows for fine-grained control over this trade-off between hallucination and detail retention.

\subsection{Additional Annotation Details}
\label{apx:human_annotation}
To recruit high-quality annotators, we required them to have a HIT rate greater than 97\% and at least 5,000 approved HITs, without imposing any geographical constraints. Additionally, we designed a qualification test consisting of 10 questions of varying difficulty, which we manually evaluated. A total of 14 annotators who passed the test with fewer than 2 mistakes were selected to perform the annotations for all experiments.

\begin{figure*}[ht]
    \centering
    \includegraphics[width=1.0\linewidth]{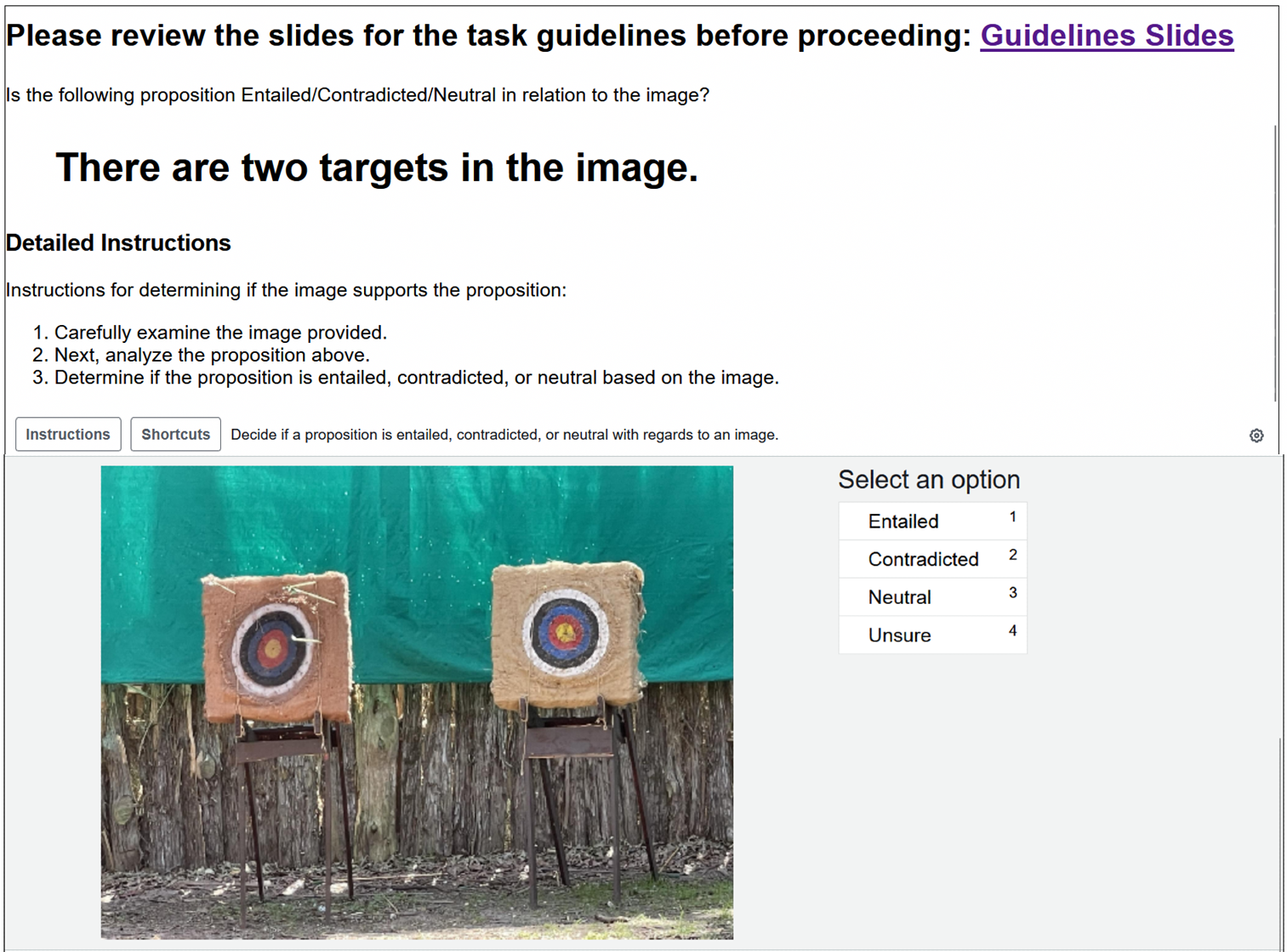}
    \vspace{-10pt}
    \caption{\textbf{Example of annotation interface in Amazon Mechanical Turk}} 
    \label{fig:amt_interface}
\end{figure*}

We present an example from the Amazon Mechanical Turk user interface, as shown to the annotators in Figure \ref{fig:amt_interface}. Alongside the generated task guidelines, we provided supplementary slides with a detailed explanation of the visual entailment task, including 10 manually annotated examples that feature different labels and varying levels of difficulty for the annotators to review.

Annotators were paid 0.10\$ per annotation, with an average hourly wage of 10\$. The annotation process was approved by the institution's ethics committee.

\subsection{Additional Experiments}
\label{sec:more_experiments}

\medskip \noindent \textbf{Larger-Scale Dense Caption Dataset}
We fine-tune PaliGemma on a dataset that is 10 times larger, sampling 100,000 image-caption pairs from the PixelProse~\cite{singla2024pixels} dataset. Of these, 95,000 pairs are used for training and 5,000 for testing. We follow the same configurations as in the main paper, except that we train for a single epoch, and report the results in Table \ref{tab:pixelprose}. As shown, \methodname{} significantly reduces the contradiction rate while maintaining high descriptiveness, demonstrating its effectiveness in generalizing to larger-scale datasets.

\medskip \noindent \textbf{Relative Location of Contradicted Propositions.}\ In Figure \ref{fig:cont_location}, we illustrate the distributions of the relative locations of propositions categorized as contradicted with respect to the generated caption. Consistent with prior research \cite{mckenna2023sources} We observe that errors tend to increase with the distance from the beginning of the text: as more text is written, the rate of contradictions increases.

\medskip \noindent \textbf{DNLI Evaluation with a Smaller Model}\
To justify the use of Gemini in the question difficulty evaluation stage, we compared its performance against Gemma-2B by using both models as judges on 1,000 question-answer pairs from the DOCCI test set. The results showed a 75\% agreement between the two models. To further analyze discrepancies, we manually inspected 20 random samples where the judgments differed, finding that in 18 out of 20 cases, Gemini provided the accurate assessment. These findings highlight a trade-off between model size and judgment accuracy. While Gemma-2B is a more lightweight alternative, we prioritized the superior accuracy and consistency offered by Gemini for our evaluations.

\medskip
\noindent \textbf{Additional \methodname{} Statistics} \
In Figure \ref{fig:questions_distribution}, we show the distribution of the number of questions generated per caption for the DOCCI training set. The data indicates that the distribution is centered around approximately 13 visual questions per caption.

In Figure \ref{fig:threshold_questions_distribution}, we demonstrate how the number of unknown questions changes when adjusting the threshold $T$. As T increases, more questions are classified as unknown, shifting the distribution.

\end{document}